\documentclass{article}

\usepackage{microtype}
\usepackage{graphicx}
\usepackage{booktabs} 
\usepackage{multirow} 
\usepackage{caption}
\usepackage{subcaption}
\usepackage{algorithm}
\usepackage[ngerman]{babel}
\usepackage{paralist, tabularx}
\usepackage{makecell}
\usepackage{colortbl}
\usepackage{tcolorbox}
\usepackage{appendix}
\usepackage{hyperref}
\usepackage{placeins}
\usepackage{mdframed}
\usepackage{listings}
\usepackage{xcolor}

\usepackage[accepted]{icml_fmwild}

\usepackage{amsmath}
\usepackage{amssymb}
\usepackage{mathtools}
\usepackage{amsthm}

\usepackage[capitalize,noabbrev]{cleveref}

\theoremstyle{plain}
\newtheorem{theorem}{Theorem}[section]

\newtheorem{lemma}[theorem]{Lemma}

\theoremstyle{definition}

\theoremstyle{remark}

\usepackage[textsize=tiny]{todonotes}

\icmltitlerunning{Strong Copyright Protection for Language Models via Adaptive Model Fusion}

\newcommand{\model}{p}
\newcommand{\domainconcepts}{\mathcal C}

\newcommand{\dataset}{\mathcal D}
\newcommand{\cpdelta}{\text{CP-}\Delta}
\newcommand{\modeli}{p^{(i)}}
\newcommand{\modelone}{p^{(1)}}
\newcommand{\modeltwo}{p^{(2)}}

\newif\ifshownotes
\shownotestrue

\definecolor{codegreen}{rgb}{0,0.6,0}
\definecolor{codegray}{rgb}{0.5,0.5,0.5}
\definecolor{codepurple}{rgb}{0.58,0,0.82}
\definecolor{backcolour}{rgb}{0.95,0.95,0.92}
\definecolor{lightgray}{RGB}{240,240,240}
\definecolor{lightblue}{RGB}{173,216,230}

\lstdefinestyle{mystyle}{
    backgroundcolor=\color{backcolour},   
    commentstyle=\color{codegreen},
    keywordstyle=\color{magenta},
    numberstyle=\tiny\color{codegray},
    stringstyle=\color{codepurple},
    basicstyle=\tiny\ttfamily,
    breakatwhitespace=false,         
    breaklines=true,                 
    captionpos=b,                    
    keepspaces=true,                 
    numbers=left,                    
    numbersep=5pt,                  
    showspaces=false,                
    showstringspaces=false,
    showtabs=false,                  
    tabsize=2,
}

\newmdenv[
    backgroundcolor=lightgray,
    linewidth=2pt,
    linecolor=lightblue,
    roundcorner=10pt,
    innerleftmargin=10pt,
    innerrightmargin=10pt,
    innertopmargin=10pt,
    innerbottommargin=10pt,
    frametitle={\textbf{Prompt:} How would you code a function in Python 3 to set data for outgoing stream},
    frametitlebackgroundcolor=lightblue!20
]{promptmdf}

\newmdenv[
    backgroundcolor=lightgray,
    linewidth=2pt,
    linecolor=lightblue,
    roundcorner=10pt,
    innerleftmargin=10pt,
    innerrightmargin=10pt,
    innertopmargin=10pt,
    innerbottommargin=10pt,
    frametitle={\textbf{Prompt:} Write a Python 3 function for reading package file as text to get name and version},
    frametitlebackgroundcolor=lightblue!20
]{promptmdftwo}

\newmdenv[
    backgroundcolor=lightgray,
    linewidth=2pt,
    linecolor=lightblue,
    roundcorner=10pt,
    innerleftmargin=10pt,
    innerrightmargin=10pt,
    innertopmargin=10pt,
    innerbottommargin=10pt,
    frametitle={\textbf{Prompt:} How would you implement a function in Python 3 that calculates the mean heart rate in beats per minute from a set of rr intervals},
    frametitlebackgroundcolor=lightblue!20
]{promptmdfthree}

\begin{document}

\twocolumn[
\icmltitle{Strong Copyright Protection for Language Models via Adaptive Model Fusion}



\icmlsetsymbol{equal}{*}
\icmlsetsymbol{authors}{†}

\begin{icmlauthorlist}
\icmlauthor{Javier Abad}{equal,eth}
\icmlauthor{Konstantin Donhauser}{equal,eth}
\icmlauthor{Francesco Pinto}{ox}
\icmlauthor{Fanny Yang}{eth}
\end{icmlauthorlist}

\icmlaffiliation{eth}{Department of Computer Science, ETH Zurich, Switzerland}
\icmlaffiliation{ox}{Department of Engineering Science, University of Oxford, UK}

\icmlcorrespondingauthor{Javier Abad}{javier.abadmartinez@ai.ethz.ch}

\icmlkeywords{Machine Learning, ICML}

\vskip 0.3in
]



\printAffiliationsAndNotice{\icmlEqualContribution} 

\begin{abstract}
The risk of language models unintentionally reproducing copyrighted material from their training data has led to the development of various protective measures. In this paper, we propose model fusion as an effective solution to safeguard against copyright infringement. In particular, we introduce Copyright-Protecting Fusion (CP-Fuse), an algorithm that adaptively combines language models to minimize the reproduction of protected materials. CP-Fuse is inspired by the recently proposed Near-Access Free (NAF) framework and additionally incorporates a desirable \emph{balancing property} that we demonstrate prevents the reproduction of memorized training data. Our results show that CP-Fuse significantly reduces the memorization of copyrighted content while maintaining high-quality text and code generation. Furthermore, we demonstrate how CP-Fuse can be integrated with other techniques for enhanced protection.
\end{abstract}

\section{Introduction}

Large Language Models (LLMs), such as GPT-4 \citep{achiam2023gpt} and Gemini \citep{team2023gemini}, have made remarkable progress in automating tasks traditionally requiring human ingenuity, including code generation and creative writing. However, these advancements also introduce the risk of LLMs reproducing copyrighted material from their training data \citep{yu2023codeiipprompt, Meeus2023DidTN, carlini2023quantifying, karamolegkou2023copyright}, posing substantial legal challenges and leading to multi-million dollar lawsuits \citep{henderson2023foundation}. 
As a result, preventing copyright infringement in language models has become a critical concern for researchers and practitioners alike.

An approach to reducing the generation of copyrighted material involves curating training data to exclude or deduplicate protected samples \citep{ kandpal2022deduplicating, ippolito2023donottrain, carlini2023quantifying}. However, this process is resource-intensive and may not be entirely effective \citep{lee2023talkin, ippolito2023preventing}.  Additionally, copyrighted samples often represent high-quality inputs crucial for the models' performance \citep{Meeus2023DidTN}, making their exclusion potentially undesirable. In fact, under the fair use doctrine (17 U.S.C. §107), leveraging protected material is permitted provided the output does not substitute the copyrighted work or harm its market \citep{henderson2023foundation, rahman2023beyond}. Therefore, practitioners seek strategies to train models with copyrighted content while preventing infringements once deployed \citep{wei2024evaluating}.

To address this issue, various works focus on mitigating the memorization phenomenon in LLMs \citep{carlini2019secret, carlini2021extracting, carlini2023quantifying, zhang2023counterfactual, nasr2023scalable}, aiming to prevent them from reproducing verbatim text from their training data. Several methods intervene \emph{during the training phase}, and propose strategies for training or fine-tuning generative models with protected data while ensuring their outputs remain copyright-compliant \citep{anil2022large, chu2024protect, hans2024like}. Although promising, these approaches are usually computationally demanding, can compromise model utility \citep{anil2022large}, or rely on heuristics without guarantees for preventing memorization \citep{hans2024like}. Other approaches explicitly permit language models full access to copyrighted material during training and intervene \emph{during the inference phase} \citep{ippolito2023preventing, vyas2023provable}. Notably, \citet{vyas2023provable} introduce a general approach for constructing copyright-protected models by fusing generative models trained on different data sources. However, while their framework shows potential, it currently lacks tractable algorithms for implementation.



In this paper, we propose Copyright-Protecting Fusion (CP-Fuse), a simple algorithm for copyright-protecting model fusion. Our method builds upon an extensive body of literature on model fusion for language models \citep{liu2021dexperts, jiang2023llm, gururangan2023scaling, wang2023fusing, mavromatis2024pack}. In particular, CP-Fuse adaptively aggregates the logits to reduce the probability of regurgitating copyrighted content.  In \Cref{sec:algo}, we introduce the algorithm, which we derive from the Near-Access Free (NAF) framework \citep{vyas2023provable}. We demonstrate that it adheres to a \textit{balancing property} (\Cref{lem:balance}), which intuitively explains how it prevents the regurgitation of copyright-protected material when using greedy decoding strategies. In \Cref{sec:experiments}, we showcase the effectiveness of CP-Fuse in preventing language models from reproducing memorized training samples, reducing regurgitation by more than $25\times$ compared to copyright-infringing models. Additionally, we show that it consistently outperforms other techniques aimed at preventing memorization during the inference phase while maintaining competitive perplexity and generating high-quality code and text. Finally, preliminary experiments indicate that our method can be combined on top of other copyright-protecting strategies, such as those intervening in the training phase, for enhanced protection.

\section{Related Works on Copyright Protection}
\label{sec:relatedwork}


Measures for copyright protection can be implemented at various stages of model deployment \citep{lee2023talkin}. Since many open-source LLMs are trained on datasets containing copyrighted material, such as the BookCorpus dataset (e.g., GPT-3 \citep{brown2020language}) and the C4 corpus (e.g., LLaMa \citep{touvron2023llama}), efforts have been made to curate datasets with exclusively licensed content \citep{kocetkov2022stack, min2023silo}. Moreover, removing duplicated copyrighted samples from the dataset has been shown to reduce regurgitation \citep{kandpal2022deduplicating}. However, these approaches are resource-intensive and can degrade model performance \citep{ippolito2023preventing, Meeus2023DidTN}.


Other approaches intervene during the training or fine-tuning of LLMs. These methods usually aim to prevent memorization, as verbatim reproduction can constitute copyright infringement in text and code \citep{yu2023codeiipprompt, henderson2023foundation}. In this context, differential privacy (DP) \citep{dwork2014algorithmic, abadi2016deep} offers a solution against memorization by limiting the influence of individual training points on the model's output. However, DP training is computationally demanding, usually degrades generation utility \citep{anil2022large}, and its goals differ from those of copyright protection \citep{elkin2023can}. Additionally, heuristic alternatives, such as the goldfish loss \citep{hans2024like} or simple early stopping \citep{mireshghallah2022memorization}, have proven effective in preventing the regurgitation of training text, though they lack theoretical guarantees.

An orthogonal line of work allows language models full access to copyrighted content during training and enforces copyright constraints via post-processing \citep{wei2024evaluating}. Filtering strategies, such as Mem-Free \citep{ippolito2023donottrain}, can effectively prevent verbatim reproduction of copyrighted material from a curated blocklist at inference time. However, these methods only work for consecutive verbatim matches, and may lead to hallucinations due to modifications in the decoding process \citep{liu2024shield}. Furthermore, several works propose unlearning copyrighted content from trained models \citep{bourtoule2021machine,chen2023unlearn, eldan2023s, jang2023knowledge, zhang2024negative, liu2024rethinking}; however, these approaches are typically computationally impractical and require access to model weights, which is restrictive in real-world scenarios.
Our method, presented in \Cref{sec:algo}, derives from the Near-Access Free (NAF) framework \citep{vyas2023provable}. Unlike purely heuristic approaches, it offers a theoretical understanding of how it prevents regurgitation (\Cref{lem:balance}), and we validate its effectiveness through extensive experiments, demonstrating its ability to prevent copyright-infringing models from reproducing protected content (\Cref{sec:experiments}).


\section{Preliminaries}\label{sec:preliminaries}

We focus on language models \( p \) that take a prompt \( x \) as input and return a probability distribution over a sequence of tokens of variable length \( T \) from a fixed alphabet \( V \), with \( y_T = \text{EOS} \) representing the end-of-sequence token. Using the convention that  \( y_{<0} = \emptyset \), we can factorize $p$ as:
$$
p(y_{0:T} | x) = \prod_{t=0}^T p(y_t | y_{<t}, x).$$
In the following, we introduce a key assumption underlying our work and motivate our copyright-protection method.

\paragraph{Separability of copyrighted material}
At the core of our method is the assumption of the \textit{separability of copyrighted material}, discussed by \citet{vyas2023provable} for various vision and language applications. This assumption is akin to those used in exact machine unlearning \citep{bourtoule2021machine, yan2022arcane, dukler2023safe, kumar2023privacy} and in works that rely on splitting datasets into safe and unsafe parts \citep{golatkar2021mixed, golatkar2024cpr, li2024purifying}.

Consider a dataset $\dataset$ and a set of copyright-protected materials $\domainconcepts$ that could be compromised when training a language model $\model$ on $\dataset$. The assumption states that we can split the training data $\dataset$ into two potentially overlapping subsets, $\dataset_1$ and $\dataset_2$, such that each subset contains data associated with two mutually exclusive sets of copyright-protected materials, $\domainconcepts_1$ and $\domainconcepts_2$, where $\domainconcepts_1 \cap \domainconcepts_2 = \emptyset$. This assumption holds, for instance, when we construct the training data $\dataset$ from multiple data sources that are sufficiently distinct. Consequently, any language model trained on the subset $\dataset_1$ is protected from infringing the copyright of materials in $\domainconcepts \setminus \domainconcepts_1 \supseteq \domainconcepts_2$.

\paragraph{Near-Access Freeness (NAF)}
Given two generative models $\model^{(1)}$ and $\model^{(2)}$ trained on $\mathcal{D}_1$ and $\mathcal{D}_2$, respectively, the challenge is to construct a model $p$ that achieves protection against all copyright-protected materials $\mathcal{C}$. In that light,\citet{vyas2023provable} propose the $k$-NAF framework as a quantitative guarantee for copyright protection. 
Formally, a model $ p(. | x)$ satisfies the $k$-NAF guarantee if, for any input prompt $x$ and   some user-specified divergence function~$\Delta$,
\begin{equation}\label{eq:minkl}
  \forall x: \quad  \max_{i \in \{1,2\}} \Delta( p(. | x) \mid\mid p^{(i)}(. | x)) \leq k.
\end{equation}
The key intuition behind Equation~\eqref{eq:minkl} is that, if the separability of copyrighted material holds, the likelihood of generating copyright-infringing text for any material $c \in \domainconcepts$ is exponentially small for at least one of the models. Thus, for a model $p$ to satisfy the $k$-NAF guarantee,  it must place minimal weight on such events.  
However, it remains unclear to what extent this intuition yields meaningful guarantees for greedy search and sampling decoding strategies.



\paragraph{Model fusion with LLMs}
Independent of copyright protection, combining multiple language models is a popular research field aimed at achieving knowledge fusion, both at inference time \citep{liu2021dexperts, jiang2023llm, gururangan2023scaling, mavromatis2024pack} and after training through the merging of learned weights \citep{wortsman2022model, jin2022dataless, hsu2024let}. Most relevant to this paper are the former approaches, which generally define a model 
$p$ at inference time by combining multiple models $p^{(1)} ,\cdots , p^{(K)}$ via a weighted sum of their logits:
\begin{equation}
\label{eq:vyas}
\log p(y_t | y_{<t}, x )  := 
    \sum_{i=1}^K \alpha_t^{(i)}(y_{<t}, x) \log p^{(i)}(. | y_{<t}, x) + c,
\end{equation}
where $c$ is a normalizing constant and $\alpha_t^{(i)}$ can depend on the prompt $x$ and the history $y_{<t}$.
However, unlike our algorithm presented in the next section, these approaches do not enforce $p$ to be close to all models $p^{(i)}$ simultaneously.

\section{Copyright-Protecting Model Fusion}
\label{sec:algo}

We present \textbf{C}opyright-\textbf{P}rotecting  \textbf{Fus}ion (CP-Fuse), a simple yet effective algorithm for copyright protection in language models via model fusion. Inspired by the $k$-NAF framework, we aim to minimize the maximum KL-divergence from Equation~\eqref{eq:minkl}. Since achieving this directly is computationally intractable, we propose an efficient approximate algorithm that iteratively optimizes for $p(y_{t} | y_{<t}, x )$ given the probability of the history
$p(y_{<t} | x )$. We show in Lemma~\ref{lem:kkt} that leveraging the KL-divergence allows us to derive an update rule in the form of Equation~\eqref{eq:vyas}, commonly used in model fusion. Formally, we iteratively define 
\begin{equation}
\label{eq:pt}
    \begin{aligned}
        p(y_t \, | \, y_{<t}, x) &= \arg\min_{p^*} \max_i \underset{y_t \sim p^*}{\mathbb{E}} \log\left( \frac{p^*(y_t) p(y_{<t} \, | \, x)}{\modeli(y_{\leq t} \, | \, x)} \right)\\
            &=\arg\min_{p^* , t} ~t \quad \text{s.t.}~~\\ \forall i: ~~ \text{KL}( p^*\vert &\vert p^{(i)}(. \vert y_{<t}, x) ) + \log\left(\frac{p(y_{<t} \, | \, x)}{\modeli(y_{< t} \, | \, x)} \right) \leq t,
    \end{aligned}
\end{equation}
which results in a convex optimization problem. While solving this problem naively is still computationally intensive, we overcome this limitation using the following lemma:
\begin{lemma}
\label{lem:kkt}
The optimal solution \(  p(y_t \, | \, y_{<t}, x)\) of the optimization problem in Equation~\eqref{eq:pt} satisfies\footnote{We set \(\log(0) = - \infty\)}
\begin{equation}
    \begin{aligned}
    \log p^*(y_t) &= \alpha_t \log \modelone(y_t | y_{<t}, x) \\ & \quad + \beta_t \log \modeltwo(y_t | y_{<t}, x) + \gamma_t
    \end{aligned}
\label{eq:opt}
\end{equation}
for some \(\alpha_t, \beta_t \geq 0, \gamma_t \in \mathbb{R}\).
\end{lemma}
Consequently, the optimization problem in Equation~\eqref{eq:pt} can be solved efficiently by performing a grid search over the parameters \(\alpha_t\) and \(\beta_t\), and selecting \(\gamma_t\) as a function of \(\alpha_t\) and \(\beta_t\) to ensure that the total mass is 1.

\subsection{Discussion}
CP-Fuse adaptively selects $\alpha_t$ and $\beta_t$ based on the sequence history $y_{<t}$. In particular, the algorithm assigns less weight to the model that has been more dominant in generating $y_{<t}$, which is key for achieving strong copyright protection. More formally, the following balancing property holds:
\begin{lemma}
\label{lem:balance}
(Balancing property) Let \( y_{<t} \) be any non-ending sequence and assume that $p^{(i)}(. \vert y_{<t} , x) $ has full support for both $i\in \{1,2\}$ and  \( p^{(1)}(y_{<t} | x) > \modeltwo(y_{<t} | x) \). Then, either of the two cases is true:
\begin{align*}
    1. &  \underset{y_{t} \sim p(. | y_{<t})}{\mathbb{E}} \log \modelone(y_{\leq t})  = \underset{y_{t} \sim p(. | y_{<t})}{\mathbb{E}} \log \modeltwo(y_{\leq t}) \\
    2. &  \quad \quad \quad \quad \quad p(y_t | y_{<t}, x)  = \modeltwo(y_t | y_{<t}, x) 
\end{align*}
\end{lemma}
This balancing property ensures that neither model dominates the text generation. As an example, suppose the generation of a subsequence \( y_{<t} \) is strongly dominated by $\modelone$, such that \( \modelone(y_{<t} | x) \gg \modeltwo(y_{<t} | x) \). If the first case in \Cref{lem:balance} holds, the output distribution of the copyright-protected model, \( p(y_t | y_{<t}, x) \), will be much closer to \( \modeltwo(y_{<t} | x) \) than to \( \modelone(y_{<t} | x) \). Conversely, if the second case holds, then \( p = \modeltwo \), and the generation of \( y_t \) will be independent of \( \modelone(y_{<t} | x) \). In other words, the next token generated by \( p \) will likely match the most probable token under the dominant model, \( \modelone(y_{<t} | x) \), only if both \( \modelone \) and \( \modeltwo \) are close conditioned on \( y_{<t} \) and \( x \), that is, when the generated sequence is not protected assuming separability of copyrighted material (\Cref{sec:preliminaries}). 
We provide experimental evidence for this property in \Cref{app:weights}.

\paragraph{Comparison with related works}  \citet{vyas2023provable} propose $\cpdelta$ as a general strategy for combining two generative models. Nevertheless, their approach becomes computationally intractable when directly applied to the probability distribution  $p(.| x)$  over the entire sequence $y_T$. To address this issue, the authors suggest applying $\cpdelta$ token-wise, resulting in the model from~\Cref{eq:opt} with $\alpha_t = \beta_t = 1/2$. This algorithm has also been used in a slightly different setting for purifying language models by \citet{li2024purifying}.  

\begin{figure}[h!]
    \centering
        \includegraphics[width=0.48\textwidth]{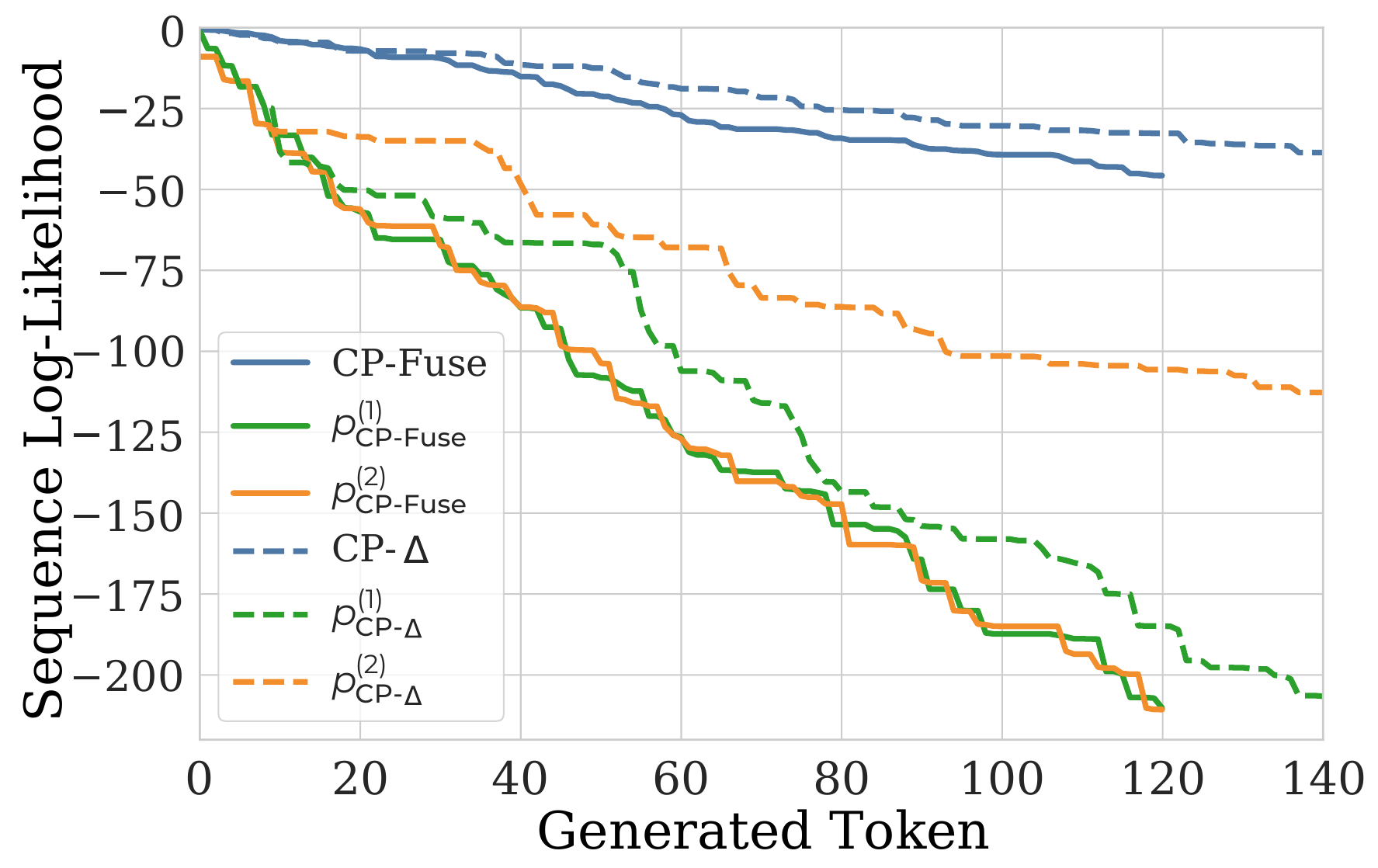}
\caption{Log-likelihood of the sequences produced by CP-Fuse and $\cpdelta$, and their base models $p^{(1)}$ and $p^{(2)}$, at each generated token. We show a random generation from StarCoder models fine-tuned on the Python instructional dataset, see \Cref{sec:experiments} for details.}
    \label{fig:cumulative_main}
\end{figure}

However,  we remark that adaptively choosing $\alpha_t$ and $\beta_t$ is crucial for achieving strong copyright protection. To illustrate this point, we present in \Cref{fig:cumulative_main} the cumulative log-likelihood at each generated token for sequences produced by CP-Fuse and $\cpdelta$, along with their respective base models $p^{(1)}$ and $p^{(2)}$. The balancing property of CP-Fuse ensures that the log-likelihood under $p^{(1)}$ and $p^{(2)}$ is approximately the same for each generated token of the sequence, thereby preventing the reproduction of copyrighted material since no protected content can be memorized by both base models. Conversely, $\cpdelta$ exhibits a clear preference towards $p^{(2)}$. This dominance suggests that $p^{(2)}$ potentially memorized training samples, making $\cpdelta$ vulnerable to reproducing them. In the next section, we further validate this observation through extensive real-world experiments.

\begin{table*}[t!]
\caption{Copyright-infringement metrics \textbf{averaged at the 95th percentile} for the Python instructions and Math abstracts datasets. We present results for the overfitted models, CP-Fuse, and $\cpdelta$. Metrics include Exact Matching (EM), Infringement Count over 160 characters (IC\textsubscript{160}), ROUGE-L (ROU), BLEU score (BLE), Levenshtein Distance (LEV). $\downarrow$ Means lower is better, $\uparrow$ means higher is better.}
\centering
\begin{tabular}{ll|ccccc|ccccc}
\toprule
& & \multicolumn{5}{c|}{\textbf{Python instructions (StarCoder)}} & \multicolumn{5}{c}{\textbf{Math abstracts (LLaMa2)}} \\
\midrule
\textbf{Model} & \textbf{Split} & \textbf{EM $\downarrow$} & \textbf{IC\textsubscript{160} $\downarrow$} & \textbf{ROU $\downarrow$} & \textbf{BLE $\downarrow$} & \textbf{LEV $\uparrow$} & \textbf{EM $\downarrow$} & \textbf{IC\textsubscript{160} $\downarrow$} & \textbf{ROU $\downarrow$} & \textbf{BLE $\downarrow$} & \textbf{LEV $\uparrow$} \\
\midrule
\multirow{3}{*}{\makecell{Overfit \\ Split 1}} & Split 1 & \multicolumn{1}{c}{2489.28} & \multicolumn{1}{c}{2329.96} & \multicolumn{1}{c}{1.00} & \multicolumn{1}{c}{1.00} & \multicolumn{1}{c|}{0.00} & \multicolumn{1}{c}{1397.68} & \multicolumn{1}{c}{1482.84} & \multicolumn{1}{c}{1.00} & \multicolumn{1}{c}{1.00} & \multicolumn{1}{c}{0.00} \\
& Split 2 & \multicolumn{1}{c}{33.74} & \multicolumn{1}{c}{0.12} & \multicolumn{1}{c}{0.54} & \multicolumn{1}{c}{0.77} & \multicolumn{1}{c|}{0.54} & \multicolumn{1}{c}{31.00} & \multicolumn{1}{c}{0.00} & \multicolumn{1}{c}{0.25} & \multicolumn{1}{c}{0.06} & \multicolumn{1}{c}{0.70} \\
& Test & \multicolumn{1}{c}{65.88} & \multicolumn{1}{c}{0.34} & \multicolumn{1}{c}{0.55} & \multicolumn{1}{c}{0.80} & \multicolumn{1}{c|}{0.49} & \multicolumn{1}{c}{28.76} & \multicolumn{1}{c}{0.00} & \multicolumn{1}{c}{0.22} & \multicolumn{1}{c}{0.15} & \multicolumn{1}{c}{0.70} \\
\midrule
\multirow{3}{*}{\makecell{Overfit \\ Split 2}} & Split 1 & \multicolumn{1}{c}{47.88} & \multicolumn{1}{c}{0.31} & \multicolumn{1}{c}{0.53} & \multicolumn{1}{c}{0.82} & \multicolumn{1}{c|}{0.52} & \multicolumn{1}{c}{42.12} & \multicolumn{1}{c}{0.00} & \multicolumn{1}{c}{0.27} & \multicolumn{1}{c}{0.08} & \multicolumn{1}{c}{0.68} \\
& Split 2 & \multicolumn{1}{c}{2182.16} & \multicolumn{1}{c}{2019.48} & \multicolumn{1}{c}{1.00} & \multicolumn{1}{c}{1.00} & \multicolumn{1}{c|}{0.00} & \multicolumn{1}{c}{1570.88} & \multicolumn{1}{c}{1688.72} & \multicolumn{1}{c}{1.00} & \multicolumn{1}{c}{1.00} & \multicolumn{1}{c}{0.00} \\
& Test & \multicolumn{1}{c}{41.38} & \multicolumn{1}{c}{0.17} & \multicolumn{1}{c}{0.53} & \multicolumn{1}{c}{0.66} & \multicolumn{1}{c|}{0.52} & \multicolumn{1}{c}{37.48} & \multicolumn{1}{c}{0.00} & \multicolumn{1}{c}{0.26} & \multicolumn{1}{c}{0.07} & \multicolumn{1}{c}{0.69} \\
\midrule
\midrule
\multirow{3}{*}{CP-Fuse} & Split 1 & \multicolumn{1}{c}{59.48} & \multicolumn{1}{c}{0.60} & \multicolumn{1}{c}{0.81} & \multicolumn{1}{c}{0.66} & \multicolumn{1}{c|}{0.23} & \multicolumn{1}{c}{94.20} & \multicolumn{1}{c}{0.00} & \multicolumn{1}{c}{0.35} & \multicolumn{1}{c}{0.14} & \multicolumn{1}{c}{0.62} \\
& Split 2 & \multicolumn{1}{c}{48.88} & \multicolumn{1}{c}{1.30} & \multicolumn{1}{c}{0.81} & \multicolumn{1}{c}{0.64} & \multicolumn{1}{c|}{0.24} & \multicolumn{1}{c}{65.84} & \multicolumn{1}{c}{0.00} & \multicolumn{1}{c}{0.34} & \multicolumn{1}{c}{0.14} & \multicolumn{1}{c}{0.63} \\
& Test & \multicolumn{1}{c}{35.59} & \multicolumn{1}{c}{0.04} & \multicolumn{1}{c}{0.57} & \multicolumn{1}{c}{0.71} & \multicolumn{1}{c|}{0.52} & \multicolumn{1}{c}{47.92} & \multicolumn{1}{c}{0.00} & \multicolumn{1}{c}{0.28} & \multicolumn{1}{c}{0.07} & \multicolumn{1}{c}{0.68} \\
\midrule
\multirow{3}{*}{$\cpdelta$} & Split 1 & \multicolumn{1}{c}{341.60} & \multicolumn{1}{c}{136.52} & \multicolumn{1}{c}{1.00} & \multicolumn{1}{c}{1.00} & \multicolumn{1}{c}{0.07} & \multicolumn{1}{c}{273.20} & \multicolumn{1}{c}{253.40} & \multicolumn{1}{c}{0.72} & \multicolumn{1}{c}{0.58} & \multicolumn{1}{c}{0.30} \\
& Split 2 & \multicolumn{1}{c}{162.80} & \multicolumn{1}{c}{152.64} & \multicolumn{1}{c}{1.00} & \multicolumn{1}{c}{1.00} & \multicolumn{1}{c|}{0.02} & \multicolumn{1}{c}{284.80} & \multicolumn{1}{c}{1.66} & \multicolumn{1}{c}{0.50} & \multicolumn{1}{c}{0.33} & \multicolumn{1}{c}{0.45} \\
& Test & \multicolumn{1}{c}{39.91} & \multicolumn{1}{c}{0.03} & \multicolumn{1}{c}{0.58} & \multicolumn{1}{c}{0.80} & \multicolumn{1}{c|}{0.51} & \multicolumn{1}{c}{57.50} & \multicolumn{1}{c}{0.00} & \multicolumn{1}{c}{0.29} & \multicolumn{1}{c}{0.07} & \multicolumn{1}{c}{0.67} \\
\bottomrule
\end{tabular}
\label{tab:results_copyright}
\end{table*}

\section{Experiments}
\label{sec:experiments}

\subsection{Experimental Setup}
We use large pre-trained language models that are commonly used in practical applications. We fine-tune the models on two different splits, each containing 3,000 samples. To assess the copyright protection capabilities of CP-Fuse, we consider an extreme case where each model overfits the splits by fine-tuning them for many epochs (50+, see Appendix~\ref{app:implementation}). Consequently, the base models strongly memorize the training data, representing a challenging scenario where they are prone to reproducing exactly any training sample, thereby infringing on copyright. For the experiments with early-stopped models, we stop fine-tuning after 2 epochs.

\paragraph{Datasets and Models}

We evaluate CP-Fuse across three scenarios. First, we fine-tune the LLaMa2 7B model \citep{touvron2023llama} on a dataset of abstracts from math papers\footnote{\hyperlink{https://huggingface.co/datasets/math-ai/AutoMathText}{AutoMathText} \citep{zhang2024autonomous}} using the title of each paper as the prompt. We conduct additional experiments using GPT-2 XL \citep{radford2019language} and Phi-2 \citep{javaheripi2023phi} on this dataset in \Cref{app:gpt_results}. Second, we fine-tune the StarCoder 7B model \citep{li2023starcoder} using an instructional dataset for Python\footnote{\hyperlink{https://huggingface.co/datasets/Nan-Do/instructional_code-search-net-python?row=0}{instructional\_code-search-net-python}}, where the prompts are natural language descriptions of tasks and the responses are Python code solutions. Finally, we fine-tune the StarCoder model on the APPS dataset\footnote{\hyperlink{https://huggingface.co/datasets/codeparrot/apps}{APPS} \citep{hendrycksapps2021}}, which also consists of natural language problems and Python solutions, and incorporates unit tests to asses the code generation quality. For this last task, the models are additionally evaluated on the MBPP\footnote{\hyperlink{https://huggingface.co/datasets/google-research-datasets/mbpp}{MBPP} \citep{austin2021program}} and the HumanEval\footnote{\hyperlink{https://huggingface.co/datasets/codeparrot/instructhumaneval}{InstructHumanEval} \citep{chen2021evaluating}} datasets, both of which also include unit tests. Note that both code and text-based tasks represent settings where copyright infringement is a concern \citep{yu2023codeiipprompt, henderson2023foundation}. The complete list of hyperparameters is included in \Cref{app:implementation}.


\paragraph{Metrics}
We use a wide range of metrics to measure copyright infringement. We report the average values of these metrics above the 95th percentile. 
This focus on percentiles addresses the legal concern of copying long text extracts in real-world applications. We present the metrics for the two fine-tuning splits and a test set comprising 500 prompts. 

To measure \textit{exact memorization}, we use the average Exact Matching (EM) length and the Infringement Count (IC) for substrings over 160 characters, a threshold that is consistent with regulations \citep{mueller2024llms}. Exact matching is widely recognized in the literature for evaluating memorization, as it clearly indicates copyright infringement in both text and code \citep{lee2021deduplicating, karamolegkou2023copyright, carlini2023quantifying, yu2023codeiipprompt, mueller2024llms}.

For \textit{approximate (non-verbatim) memorization}, we use the ROUGE-L, BLEU score, and normalized Levenshtein distance, consistent with the literature \citep{ippolito2023preventing, huang2023privacy, chen2024copybench}. Additional metrics and their corresponding results are included in \Cref{app:more_copyright}. We give implementation details in \Cref{app:metrics}.


We use perplexity as a utility metric for the generated code and text. For evaluating code, the APPS, MBPP, and HumanEval datasets further incorporate unit tests, allowing for the computation of the pass@1 score \citep{chen2021evaluating}.


\paragraph{Baseline} 
We compare our method against $\cpdelta$ \citep{vyas2023provable}, using the KL divergence as the divergence $\Delta$. For CP-Fuse, we construct the grid by uniformly discretizing the interval $[0, 2)$ with 10 steps and $[2, 10]$ with 9 steps.

\subsection{Results}

\begin{figure*}[t!]
    \centering
    \begin{subfigure}[b]{0.48\textwidth}
        \centering
        \includegraphics[width=\textwidth]{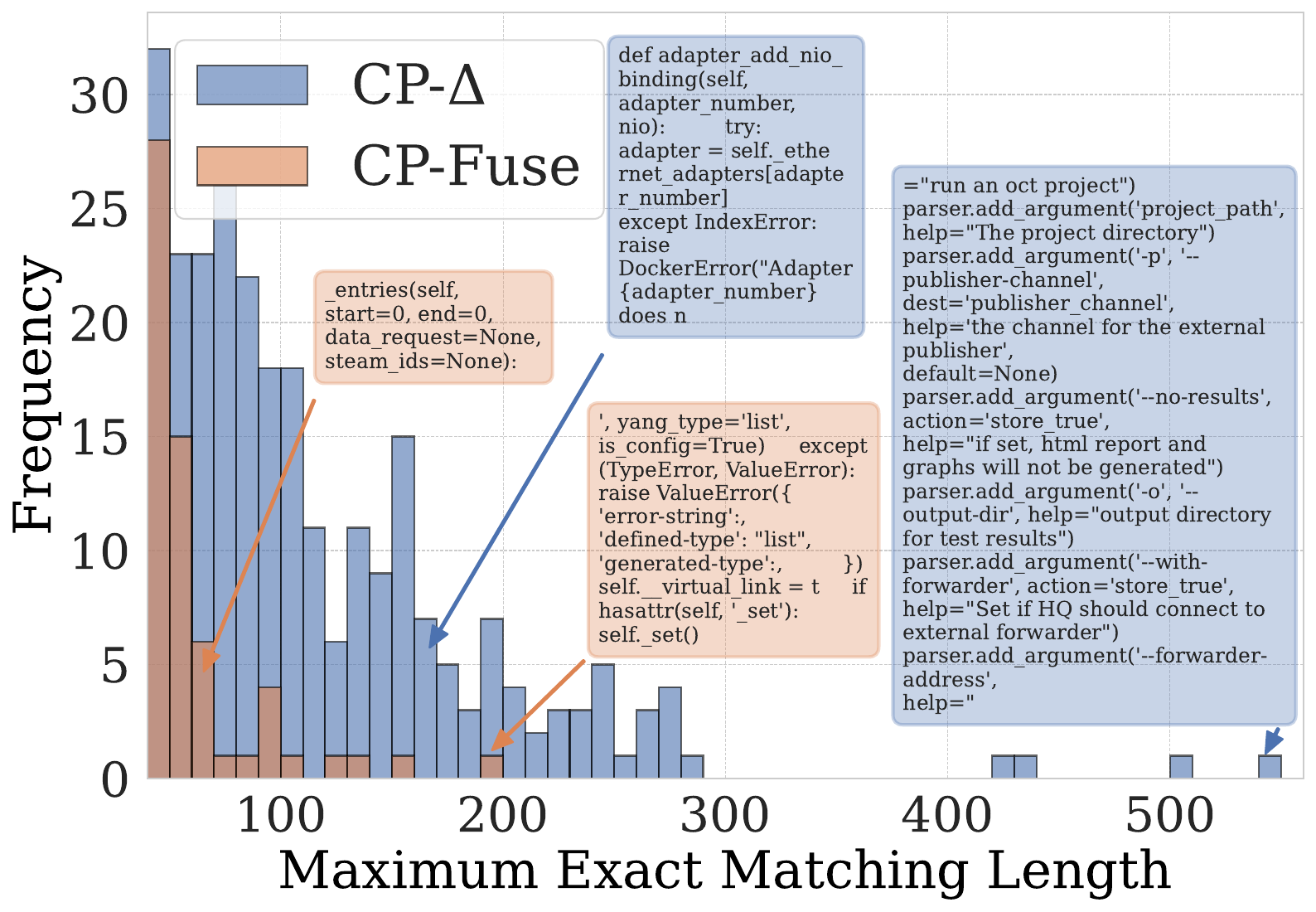}
        \caption{Python instructions (StarCoder)}
        \label{fig:python_starcoder}
    \end{subfigure}%
    \begin{subfigure}[b]{0.48\textwidth}
        \centering
        \includegraphics[width=\textwidth]{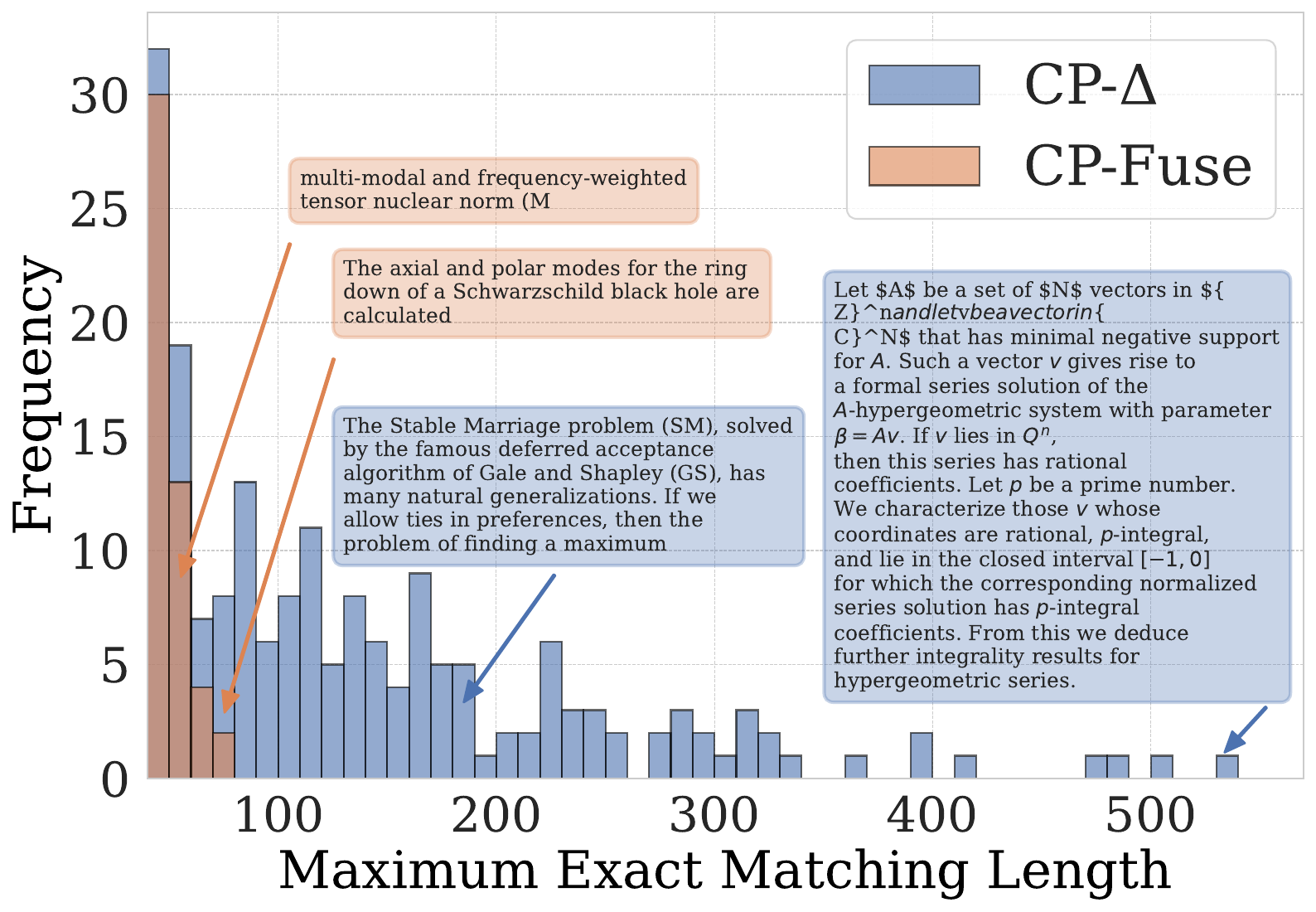}
        \caption{Math abstracts (LLaMa2)}
        \label{fig:abstracts_LLaMa}
    \end{subfigure}
    \caption{Histogram of exactly matched substring lengths (above 40 characters) generated by CP-$\Delta$ and CP-Fuse for (a) the Python instructions and (b) the math abstracts datasets. We show the longest substring and one randomly sampled match above 40 characters.}
    \label{fig:combined}
    
\end{figure*}

In this section, we present a systematic evaluation of our algorithm. In particular, we show its effectiveness in generating high-quality text and correct code while preventing the reproduction of large segments from the training data.

\paragraph{CP-Fuse significantly reduces exact and approximate memorization} 
\Cref{tab:results_copyright} shows the copyright-infringement metrics for the overfitted (copyright-infringing) models, CP-Fuse, and $\cpdelta$. From the exact memorization metrics, EM and IC\textsubscript{160}, it is evident that CP-Fuse substantially reduces regurgitation in both the code and text tasks. Specifically, CP-Fuse decreases exact matches by more than a factor of 25 compared to the overfitted models. CP-Fuse also consistently outperforms $\cpdelta$, which produces long text segments that exactly match the training data and are 3 to 6 times longer than those produced by our method. These segments often exceed the legal 160-character threshold, while such infringements almost completely vanish with our method. 

The approximate memorization metrics further support these observations, showcasing a clear and consistent improvement of CP-Fuse compared to the overfitted models and a better performance than $\cpdelta$. Additionally, our method performs closely to the overfitted models on the test set in all metrics. These findings demonstrate the effectiveness of our method in preventing both verbatim and quasi-verbatim reproduction of training material. We refer to \Cref{app:more_copyright} for additional metrics that align with these conclusions.

\begin{table}
\caption{Utility metrics for the methods. We include pass@1 for APPS, MBPP, and HumanEval (HE), and perplexity (PPL) for Math abstracts. $\downarrow$ Means lower is better, $\uparrow$ means higher is better.}
\centering
\begin{tabular}{l|ccc|c}
\toprule
& \multicolumn{3}{c|}{\textbf{pass@1} $\uparrow$} &  \textbf{PPL} $\downarrow$  \\
\midrule
\textbf{Metric} & \textbf{APPS} & \textbf{MBPP} & \textbf{HE} & \textbf{Math Abs.} \\
\midrule
\makecell{Overfit Split 1} & 0.43 & 0.44 & 0.29 & 1.41 \\
\makecell{Overfit Split 2} & 0.42 & 0.44 & 0.28 & 1.23 \\
CP-Fuse & 0.47 & 0.43 & 0.28 & 1.61 \\
$\cpdelta$ & 0.45 & 0.46 & 0.29 & 1.54 \\
\bottomrule
\end{tabular}
\label{tab:results_utility}
\end{table}

We now provide a more detailed comparison between our method and $\cpdelta$. \Cref{fig:combined} displays histograms illustrating the distribution of exact matches generated by both methods. We observe a considerably more heavy-tailed distribution for $\cpdelta$, which consistently reproduces longer verbatim text segments than CP-Fuse and hence is more likely to infringe on copyright. For example, the longest exact match for CP-Fuse in the abstracts task is 73 characters, while the 95th percentile for $\cpdelta$ is 342 characters, with the longest match exceeding 500 characters. Our results underscore the importance of adaptively setting the weights when combining the language models for effective copyright protection.

\paragraph{CP-Fuse produces high-quality code and text} \Cref{tab:results_utility}  presents the utility metrics: pass@1, and the perplexity score. For the code generation task, CP-Fuse demonstrates performance that is competitive with the overfitted models and $\cpdelta$, passing a similar proportion of unit tests. Our method also achieves low perplexity in the text generation task, comparable to the overfitted models. Note that both the overfitted models and $\cpdelta$ memorize training samples and thus generate very low-perplexity text through regurgitation. 

\begin{table}[t!]
\centering
\caption{Perplexity (PPL) and Exact Matching (EM) at the 95th percentile for StarCoder and LLaMa2. We report results for the early-stopped (ES) models, the baseline $\cpdelta$, and CP-Fuse.}
\begin{tabular}{ll|cc|cc}
\toprule
& & \multicolumn{2}{c|}{\textbf{StarCoder}} & \multicolumn{2}{c}{\textbf{LLaMa2}} \\
\textbf{Model} & \textbf{Split} & \textbf{PPL} & \textbf{EM} & \textbf{PPL} & \textbf{EM} \\
\midrule
\multirow{3}{*}{\makecell{ES Split 1}} & Split 1 & 1.26 & 159.36 & 1.46 & 207.44 \\
& Split 2 & 1.30 & 39.23 & 1.50 & 46.87 \\
& Test & 1.30 & 51.71 & 1.52 & 44.83 \\
\midrule
\multirow{3}{*}{\makecell{ES Split 2}} & Split 1 & 1.25 & 31.96 & 1.49 & 44.76 \\
& Split 2 & 1.24 & 145.04 & 1.40 & 280.20 \\
& Test & 1.27 & 43.74 & 1.47 & 44.65 \\
\midrule
\midrule
\multirow{3}{*}{CP-Fuse} & Split 1 & 1.29 & 46.96 & 1.52 & 52.21 \\
& Split 2 & 1.29 & 44.50 & 1.52 & 53.30 \\
& Test & 1.29 & 49.43 & 1.53 & 45.00 \\
\midrule
\multirow{3}{*}{CP-$\Delta$} & Split 1 & 1.29 & 70.17 & 1.46 & 68.84 \\
& Split 2 & 1.29 & 59.04 & 1.45 & 61.48 \\
& Test & 1.30 & 48.12 & 1.46 & 45.79 \\
\bottomrule
\end{tabular}
\label{tab:wrapping_results}
\end{table}

We further validate the high quality of text and code produced by our method with extracts of its generated outputs in \Cref{app:examples}. For the code task, CP-Fuse produces significantly different code from the original, effectively solving the task and often incorporating exception handling and additional features. Furthermore, for the text task, it generates reasonable and coherent abstracts from the paper titles.


\paragraph{CP-Fuse can be used on top of any model for enhanced copyright protection} 
Finally, we emphasize that CP-Fuse can be applied on top of any black-box model to reduce copyright infringement. For instance, we demonstrate how CP-Fuse can be combined with other measures, such as early stopping, a straightforward method for mitigating memorization issues. As shown in \Cref{tab:wrapping_results}, CP-Fuse reduces the regurgitation of memorized training samples compared to using early stopping alone. Notably, it reduces regurgitation by a factor of 3 for StarCoder and by a factor of 4 for LLaMa2, consistently outperforming the baseline $\cpdelta$.




\section{Conclusions}

In this paper, we introduced CP-Fuse, a simple yet highly effective algorithm for copyright protection based on model fusion. We first demonstrated that CP-Fuse satisfies desirable properties for preventing the reproduction of memorized samples. Moreover, we presented evidence of its effectiveness in challenging scenarios with heavily overfitted, copyright-infringing models, where CP-Fuse significantly reduced memorization without compromising the quality of generated content. The versatility of CP-Fuse was also illustrated by its seamless integration with other techniques like early stopping to further mitigate memorization issues.

An avenue for future research is exploring how CP-Fuse performs when the separability of copyrighted material partially holds. Additionally, a more nuanced evaluation of its problem-solving capabilities would be fruitful, though assessing the utility of LLMs remains inherently challenging. Finally, we suggest future work to evaluate CP-Fuse as a wrapper for other mitigation methods, such as the goldfish loss, or fine-tuning strategies like low-rank adaptations.



\section*{Acknowledgements}

JA was supported by the ETH AI Center. KD was supported by the ETH AI Center and the ETH Foundations of Data Science.

\bibliography{icml}

\begin{thebibliography}{63}
\providecommand{\natexlab}[1]{#1}
\providecommand{\url}[1]{\texttt{#1}}
\expandafter\ifx\csname urlstyle\endcsname\relax
  \providecommand{\doi}[1]{doi: #1}\else
  \providecommand{\doi}{doi: \begingroup \urlstyle{rm}\Url}\fi

\bibitem[Abadi et~al.(2016)Abadi, Chu, Goodfellow, McMahan, Mironov, Talwar, and Zhang]{abadi2016deep}
Abadi, M., Chu, A., Goodfellow, I., McMahan, H.~B., Mironov, I., Talwar, K., and Zhang, L.
\newblock Deep learning with differential privacy.
\newblock In \emph{Proceedings of the 2016 ACM SIGSAC conference on computer and communications security}, pp.\  308--318, 2016.

\bibitem[Achiam et~al.(2023)Achiam, Adler, Agarwal, Ahmad, Akkaya, Aleman, Almeida, Altenschmidt, Altman, Anadkat, et~al.]{achiam2023gpt}
Achiam, J., Adler, S., Agarwal, S., Ahmad, L., Akkaya, I., Aleman, F.~L., Almeida, D., Altenschmidt, J., Altman, S., Anadkat, S., et~al.
\newblock {GPT}-4 technical report.
\newblock \emph{arXiv preprint arXiv:2303.08774}, 2023.

\bibitem[Anil et~al.(2022)Anil, Ghazi, Gupta, Kumar, and Manurangsi]{anil2022large}
Anil, R., Ghazi, B., Gupta, V., Kumar, R., and Manurangsi, P.
\newblock Large-scale differentially private {BERT}.
\newblock In \emph{Findings of the Association for Computational Linguistics: EMNLP 2022}, pp.\  6481--6491, 2022.

\bibitem[Austin et~al.(2021)Austin, Odena, Nye, Bosma, Michalewski, Dohan, Jiang, Cai, Terry, Le, et~al.]{austin2021program}
Austin, J., Odena, A., Nye, M., Bosma, M., Michalewski, H., Dohan, D., Jiang, E., Cai, C., Terry, M., Le, Q., et~al.
\newblock Program synthesis with large language models.
\newblock \emph{arXiv preprint arXiv:2108.07732}, 2021.

\bibitem[Bourtoule et~al.(2021)Bourtoule, Chandrasekaran, Choquette-Choo, Jia, Travers, Zhang, Lie, and Papernot]{bourtoule2021machine}
Bourtoule, L., Chandrasekaran, V., Choquette-Choo, C.~A., Jia, H., Travers, A., Zhang, B., Lie, D., and Papernot, N.
\newblock Machine unlearning.
\newblock In \emph{2021 IEEE Symposium on Security and Privacy (SP)}, pp.\  141--159. IEEE, 2021.

\bibitem[Brown et~al.(2020)Brown, Mann, Ryder, Subbiah, Kaplan, Dhariwal, Neelakantan, Shyam, Sastry, Askell, et~al.]{brown2020language}
Brown, T., Mann, B., Ryder, N., Subbiah, M., Kaplan, J.~D., Dhariwal, P., Neelakantan, A., Shyam, P., Sastry, G., Askell, A., et~al.
\newblock Language models are few-shot learners.
\newblock \emph{Advances in neural information processing systems}, 33:\penalty0 1877--1901, 2020.

\bibitem[Carlini et~al.(2019)Carlini, Liu, Erlingsson, Kos, and Song]{carlini2019secret}
Carlini, N., Liu, C., Erlingsson, {\'U}., Kos, J., and Song, D.
\newblock The secret sharer: Evaluating and testing unintended memorization in neural networks.
\newblock In \emph{28th USENIX security symposium (USENIX security 19)}, pp.\  267--284, 2019.

\bibitem[Carlini et~al.(2021)Carlini, Tramer, Wallace, Jagielski, Herbert-Voss, Lee, Roberts, Brown, Song, Erlingsson, et~al.]{carlini2021extracting}
Carlini, N., Tramer, F., Wallace, E., Jagielski, M., Herbert-Voss, A., Lee, K., Roberts, A., Brown, T., Song, D., Erlingsson, U., et~al.
\newblock Extracting training data from large language models.
\newblock In \emph{30th USENIX Security Symposium (USENIX Security 21)}, pp.\  2633--2650, 2021.

\bibitem[Carlini et~al.(2023)Carlini, Ippolito, Jagielski, Lee, Tram{\`e}r, and Zhang]{carlini2023quantifying}
Carlini, N., Ippolito, D., Jagielski, M., Lee, K., Tram{\`e}r, F., and Zhang, C.
\newblock Quantifying memorization across neural language models.
\newblock In \emph{The Eleventh International Conference on Learning Representations}. OpenReview, 2023.

\bibitem[Chen \& Yang(2023)Chen and Yang]{chen2023unlearn}
Chen, J. and Yang, D.
\newblock Unlearn what you want to forget: Efficient unlearning for {LLM}s.
\newblock In \emph{The 2023 Conference on Empirical Methods in Natural Language Processing}, 2023.

\bibitem[Chen et~al.(2021)Chen, Tworek, Jun, Yuan, de~Oliveira~Pinto, Kaplan, Edwards, Burda, Joseph, Brockman, Ray, Puri, Krueger, Petrov, Khlaaf, Sastry, Mishkin, Chan, Gray, Ryder, Pavlov, Power, Kaiser, Bavarian, Winter, Tillet, Such, Cummings, Plappert, Chantzis, Barnes, Herbert-Voss, Guss, Nichol, Paino, Tezak, Tang, Babuschkin, Balaji, Jain, Saunders, Hesse, Carr, Leike, Achiam, Misra, Morikawa, Radford, Knight, Brundage, Murati, Mayer, Welinder, McGrew, Amodei, McCandlish, Sutskever, and Zaremba]{chen2021evaluating}
Chen, M., Tworek, J., Jun, H., Yuan, Q., de~Oliveira~Pinto, H.~P., Kaplan, J., Edwards, H., Burda, Y., Joseph, N., Brockman, G., Ray, A., Puri, R., Krueger, G., Petrov, M., Khlaaf, H., Sastry, G., Mishkin, P., Chan, B., Gray, S., Ryder, N., Pavlov, M., Power, A., Kaiser, L., Bavarian, M., Winter, C., Tillet, P., Such, F.~P., Cummings, D., Plappert, M., Chantzis, F., Barnes, E., Herbert-Voss, A., Guss, W.~H., Nichol, A., Paino, A., Tezak, N., Tang, J., Babuschkin, I., Balaji, S., Jain, S., Saunders, W., Hesse, C., Carr, A.~N., Leike, J., Achiam, J., Misra, V., Morikawa, E., Radford, A., Knight, M., Brundage, M., Murati, M., Mayer, K., Welinder, P., McGrew, B., Amodei, D., McCandlish, S., Sutskever, I., and Zaremba, W.
\newblock Evaluating large language models trained on code, 2021.

\bibitem[Chen et~al.(2024)Chen, Asai, Mireshghallah, Min, Grimmelmann, Choi, Hajishirzi, Zettlemoyer, and Koh]{chen2024copybench}
Chen, T., Asai, A., Mireshghallah, N., Min, S., Grimmelmann, J., Choi, Y., Hajishirzi, H., Zettlemoyer, L., and Koh, P.~W.
\newblock {CopyBench}: Measuring literal and non-literal reproduction of copyright-protected text in language model generation.
\newblock \emph{arXiv preprint arXiv:2407.07087}, 2024.

\bibitem[Chu et~al.(2024)Chu, Song, and Yang]{chu2024protect}
Chu, T., Song, Z., and Yang, C.
\newblock How to protect copyright data in optimization of large language models?
\newblock In \emph{Proceedings of the AAAI Conference on Artificial Intelligence}, volume~38, pp.\  17871--17879, 2024.

\bibitem[Dukler et~al.(2023)Dukler, Bowman, Achille, Golatkar, Swaminathan, and Soatto]{dukler2023safe}
Dukler, Y., Bowman, B., Achille, A., Golatkar, A., Swaminathan, A., and Soatto, S.
\newblock Safe: Machine unlearning with shard graphs.
\newblock In \emph{Proceedings of the IEEE/CVF International Conference on Computer Vision}, pp.\  17108--17118, 2023.

\bibitem[Dwork et~al.(2014)Dwork, Roth, et~al.]{dwork2014algorithmic}
Dwork, C., Roth, A., et~al.
\newblock The algorithmic foundations of differential privacy.
\newblock \emph{Foundations and Trends{\textregistered} in Theoretical Computer Science}, 9\penalty0 (3--4):\penalty0 211--407, 2014.

\bibitem[Eldan \& Russinovich(2023)Eldan and Russinovich]{eldan2023s}
Eldan, R. and Russinovich, M.
\newblock Who's {Harry Potter? approximate unlearning in LLMs}.
\newblock \emph{arXiv preprint arXiv:2310.02238}, 2023.

\bibitem[Elkin-Koren et~al.(2023)Elkin-Koren, Hacohen, Livni, and Moran]{elkin2023can}
Elkin-Koren, N., Hacohen, U., Livni, R., and Moran, S.
\newblock Can copyright be reduced to privacy?
\newblock \emph{arXiv preprint arXiv:2305.14822}, 2023.

\bibitem[Golatkar et~al.(2021)Golatkar, Achille, Ravichandran, Polito, and Soatto]{golatkar2021mixed}
Golatkar, A., Achille, A., Ravichandran, A., Polito, M., and Soatto, S.
\newblock Mixed-privacy forgetting in deep networks.
\newblock In \emph{Proceedings of the IEEE/CVF conference on computer vision and pattern recognition}, pp.\  792--801, 2021.

\bibitem[Golatkar et~al.(2024)Golatkar, Achille, Zancato, Wang, Swaminathan, and Soatto]{golatkar2024cpr}
Golatkar, A., Achille, A., Zancato, L., Wang, Y.-X., Swaminathan, A., and Soatto, S.
\newblock {CPR}: Retrieval augmented generation for copyright protection.
\newblock In \emph{Proceedings of the IEEE/CVF Conference on Computer Vision and Pattern Recognition}, pp.\  12374--12384, 2024.

\bibitem[Gururangan et~al.(2023)Gururangan, Li, Lewis, Shi, Althoff, Smith, and Zettlemoyer]{gururangan2023scaling}
Gururangan, S., Li, M., Lewis, M., Shi, W., Althoff, T., Smith, N.~A., and Zettlemoyer, L.
\newblock Scaling expert language models with unsupervised domain discovery.
\newblock \emph{arXiv preprint arXiv:2303.14177}, 2023.

\bibitem[Hans et~al.(2024)Hans, Wen, Jain, Kirchenbauer, Kazemi, Singhania, Singh, Somepalli, Geiping, Bhatele, et~al.]{hans2024like}
Hans, A., Wen, Y., Jain, N., Kirchenbauer, J., Kazemi, H., Singhania, P., Singh, S., Somepalli, G., Geiping, J., Bhatele, A., et~al.
\newblock Be like a goldfish, don't memorize! mitigating memorization in generative {LLMs}.
\newblock \emph{arXiv preprint arXiv:2406.10209}, 2024.

\bibitem[Henderson et~al.(2023)Henderson, Li, Jurafsky, Hashimoto, Lemley, and Liang]{henderson2023foundation}
Henderson, P., Li, X., Jurafsky, D., Hashimoto, T., Lemley, M.~A., and Liang, P.
\newblock Foundation models and fair use.
\newblock \emph{arXiv preprint arXiv:2303.15715}, 2023.

\bibitem[Hendrycks et~al.(2021)Hendrycks, Basart, Kadavath, Mazeika, Arora, Guo, Burns, Puranik, He, Song, and Steinhardt]{hendrycksapps2021}
Hendrycks, D., Basart, S., Kadavath, S., Mazeika, M., Arora, A., Guo, E., Burns, C., Puranik, S., He, H., Song, D., and Steinhardt, J.
\newblock Measuring coding challenge competence with apps.
\newblock \emph{NeurIPS}, 2021.

\bibitem[Hsu et~al.(2024)Hsu, Chen, Liao, Ho, Wang, Hsu, and Shiu]{hsu2024let}
Hsu, C.-J., Chen, Y.-C., Liao, F.-T., Ho, P.-C., Wang, Y.-H., Hsu, P.-C., and Shiu, D.-s.
\newblock Let's fuse step by step: A generative fusion decoding algorithm with {LLMs} for multi-modal text recognition.
\newblock \emph{arXiv preprint arXiv:2405.14259}, 2024.

\bibitem[Huang et~al.(2023)Huang, Gupta, Zhong, Li, and Chen]{huang2023privacy}
Huang, Y., Gupta, S., Zhong, Z., Li, K., and Chen, D.
\newblock Privacy implications of retrieval-based language models.
\newblock In \emph{Proceedings of the 2023 Conference on Empirical Methods in Natural Language Processing}. Association for Computational Linguistics, 2023.

\bibitem[Ippolito \& Yu(2023)Ippolito and Yu]{ippolito2023donottrain}
Ippolito, D. and Yu, Y.~W.
\newblock {DONOTTRAIN}: A metadata standard for indicating consent for machine learning.
\newblock In \emph{Proceedings of the 40th International Conference on Machine Learning}, 2023.

\bibitem[Ippolito et~al.(2023)Ippolito, Tram{\`e}r, Nasr, Zhang, Jagielski, Lee, Choquette-Choo, and Carlini]{ippolito2023preventing}
Ippolito, D., Tram{\`e}r, F., Nasr, M., Zhang, C., Jagielski, M., Lee, K., Choquette-Choo, C.~A., and Carlini, N.
\newblock Preventing generation of verbatim memorization in language models gives a false sense of privacy.
\newblock In \emph{Proceedings of the 16th International Natural Language Generation Conference}, pp.\  28--53. Association for Computational Linguistics, 2023.

\bibitem[Jain et~al.(2023)Jain, Chiang, Wen, Kirchenbauer, Chu, Somepalli, Bartoldson, Kailkhura, Schwarzschild, Saha, et~al.]{jain2023neftune}
Jain, N., Chiang, P.-y., Wen, Y., Kirchenbauer, J., Chu, H.-M., Somepalli, G., Bartoldson, B.~R., Kailkhura, B., Schwarzschild, A., Saha, A., et~al.
\newblock {NEFTune}: Noisy embeddings improve instruction finetuning.
\newblock In \emph{The Twelfth International Conference on Learning Representations}, 2023.

\bibitem[Jang et~al.(2023)Jang, Yoon, Yang, Cha, Lee, Logeswaran, and Seo]{jang2023knowledge}
Jang, J., Yoon, D., Yang, S., Cha, S., Lee, M., Logeswaran, L., and Seo, M.
\newblock Knowledge unlearning for mitigating privacy risks in language models.
\newblock In \emph{Proceedings of the 61st Annual Meeting of the Association for Computational Linguistics (Volume 1: Long Papers)}, pp.\  14389--14408, 2023.

\bibitem[Javaheripi et~al.(2023)Javaheripi, Bubeck, Abdin, Aneja, Bubeck, Mendes, Chen, Del~Giorno, Eldan, Gopi, et~al.]{javaheripi2023phi}
Javaheripi, M., Bubeck, S., Abdin, M., Aneja, J., Bubeck, S., Mendes, C. C.~T., Chen, W., Del~Giorno, A., Eldan, R., Gopi, S., et~al.
\newblock Phi-2: The surprising power of small language models.
\newblock \emph{Microsoft Research Blog}, 2023.

\bibitem[Jiang et~al.(2023)Jiang, Ren, and Lin]{jiang2023llm}
Jiang, D., Ren, X., and Lin, B.~Y.
\newblock {LLM-Blender}: Ensembling large language models with pairwise ranking and generative fusion.
\newblock In \emph{Proceedings of the 61st Annual Meeting of the Association for Computational Linguistics (Volume 1: Long Papers)}, pp.\  14165--14178, 2023.

\bibitem[Jin et~al.(2022)Jin, Ren, Preotiuc-Pietro, and Cheng]{jin2022dataless}
Jin, X., Ren, X., Preotiuc-Pietro, D., and Cheng, P.
\newblock Dataless knowledge fusion by merging weights of language models.
\newblock In \emph{The Eleventh International Conference on Learning Representations}, 2022.

\bibitem[Kandpal et~al.(2022)Kandpal, Wallace, and Raffel]{kandpal2022deduplicating}
Kandpal, N., Wallace, E., and Raffel, C.
\newblock Deduplicating training data mitigates privacy risks in language models.
\newblock In \emph{International Conference on Machine Learning}, pp.\  10697--10707. PMLR, 2022.

\bibitem[Karamolegkou et~al.(2023)Karamolegkou, Li, Zhou, and S{\o}gaard]{karamolegkou2023copyright}
Karamolegkou, A., Li, J., Zhou, L., and S{\o}gaard, A.
\newblock Copyright violations and large language models.
\newblock In \emph{Proceedings of the 2023 Conference on Empirical Methods in Natural Language Processing}, pp.\  7403--7412, 2023.

\bibitem[Kocetkov et~al.(2022)Kocetkov, Li, Allal, Li, Mou, Ferrandis, Jernite, Mitchell, Hughes, Wolf, et~al.]{kocetkov2022stack}
Kocetkov, D., Li, R., Allal, L.~B., Li, J., Mou, C., Ferrandis, C.~M., Jernite, Y., Mitchell, M., Hughes, S., Wolf, T., et~al.
\newblock The stack: 3 {TB} of permissively licensed source code.
\newblock \emph{arXiv preprint arXiv:2211.15533}, 2022.

\bibitem[Kumar et~al.(2023)Kumar, Gangadharaiah, and Roth]{kumar2023privacy}
Kumar, V.~B., Gangadharaiah, R., and Roth, D.
\newblock Privacy adhering machine un-learning in {NLP}.
\newblock In \emph{Findings of the Association for Computational Linguistics: IJCNLP-AACL 2023 (Findings)}, pp.\  268--277, 2023.

\bibitem[Lee et~al.(2022)Lee, Ippolito, Nystrom, Zhang, Eck, Callison-Burch, and Carlini]{lee2021deduplicating}
Lee, K., Ippolito, D., Nystrom, A., Zhang, C., Eck, D., Callison-Burch, C., and Carlini, N.
\newblock Deduplicating training data makes language models better.
\newblock In \emph{Proceedings of the 60th Annual Meeting of the Association for Computational Linguistics (Volume 1: Long Papers)}, pp.\  8424--8445, 2022.

\bibitem[Lee et~al.(2023)Lee, Cooper, and Grimmelmann]{lee2023talkin}
Lee, K., Cooper, A.~F., and Grimmelmann, J.
\newblock Talkin''bout {AI} generation: Copyright and the generative-{AI} supply chain.
\newblock \emph{arXiv preprint arXiv:2309.08133}, 2023.

\bibitem[Li et~al.(2023)Li, Allal, Zi, Muennighoff, Kocetkov, Mou, Marone, Akiki, Li, Chim, et~al.]{li2023starcoder}
Li, R., Allal, L.~B., Zi, Y., Muennighoff, N., Kocetkov, D., Mou, C., Marone, M., Akiki, C., Li, J., Chim, J., et~al.
\newblock {StarCoder}: may the source be with you!
\newblock \emph{arXiv preprint arXiv:2305.06161}, 2023.

\bibitem[Li et~al.(2024)Li, Liu, Pang, Du, Guo, Liu, and Lin]{li2024purifying}
Li, T., Liu, Q., Pang, T., Du, C., Guo, Q., Liu, Y., and Lin, M.
\newblock Purifying large language models by ensembling a small language model.
\newblock \emph{arXiv preprint arXiv:2402.14845}, 2024.

\bibitem[Liu et~al.(2021)Liu, Sap, Lu, Swayamdipta, Bhagavatula, Smith, and Choi]{liu2021dexperts}
Liu, A., Sap, M., Lu, X., Swayamdipta, S., Bhagavatula, C., Smith, N.~A., and Choi, Y.
\newblock {DExperts}: Decoding-time controlled text generation with experts and anti-experts.
\newblock In \emph{Proceedings of the 59th Annual Meeting of the Association for Computational Linguistics and the 11th International Joint Conference on Natural Language Processing (Volume 1: Long Papers)}, 2021.

\bibitem[Liu et~al.(2024{\natexlab{a}})Liu, Yao, Jia, Casper, Baracaldo, Hase, Xu, Yao, Li, Varshney, et~al.]{liu2024rethinking}
Liu, S., Yao, Y., Jia, J., Casper, S., Baracaldo, N., Hase, P., Xu, X., Yao, Y., Li, H., Varshney, K.~R., et~al.
\newblock Rethinking machine unlearning for large language models.
\newblock \emph{arXiv preprint arXiv:2402.08787}, 2024{\natexlab{a}}.

\bibitem[Liu et~al.(2024{\natexlab{b}})Liu, Sun, Xu, Wu, Wang, Wang, and Gao]{liu2024shield}
Liu, X., Sun, T., Xu, T., Wu, F., Wang, C., Wang, X., and Gao, J.
\newblock {SHIELD}: Evaluation and defense strategies for copyright compliance in llm text generation.
\newblock \emph{arXiv preprint arXiv:2406.12975}, 2024{\natexlab{b}}.

\bibitem[Mavromatis et~al.(2024)Mavromatis, Karypis, and Karypis]{mavromatis2024pack}
Mavromatis, C., Karypis, P., and Karypis, G.
\newblock Pack of {LLMs}: Model fusion at test-time via perplexity optimization.
\newblock \emph{arXiv preprint arXiv:2404.11531}, 2024.

\bibitem[Meeus et~al.(2023)Meeus, Jain, Rei, and de~Montjoye]{Meeus2023DidTN}
Meeus, M., Jain, S., Rei, M., and de~Montjoye, Y.-A.
\newblock Did the neurons read your book? document-level membership inference for large language models.
\newblock \emph{ArXiv}, abs/2310.15007, 2023.
\newblock URL \url{https://api.semanticscholar.org/CorpusID:264591425}.

\bibitem[Min et~al.(2023)Min, Gururangan, Wallace, Shi, Hajishirzi, Smith, and Zettlemoyer]{min2023silo}
Min, S., Gururangan, S., Wallace, E., Shi, W., Hajishirzi, H., Smith, N.~A., and Zettlemoyer, L.
\newblock Silo language models: Isolating legal risk in a nonparametric datastore.
\newblock In \emph{The Twelfth International Conference on Learning Representations}, 2023.

\bibitem[Mireshghallah et~al.(2022)Mireshghallah, Uniyal, Wang, Evans, and Berg-Kirkpatrick]{mireshghallah2022memorization}
Mireshghallah, F., Uniyal, A., Wang, T., Evans, D., and Berg-Kirkpatrick, T.
\newblock Memorization in {NLP} fine-tuning methods.
\newblock In \emph{First Workshop on Pre-training: Perspectives, Pitfalls, and Paths Forward at ICML}, 2022.

\bibitem[Mueller et~al.(2024)Mueller, G{\"o}rge, Bernzen, Pirk, and Poretschkin]{mueller2024llms}
Mueller, F.~B., G{\"o}rge, R., Bernzen, A.~K., Pirk, J.~C., and Poretschkin, M.
\newblock {LLMs} and memorization: On quality and specificity of copyright compliance.
\newblock \emph{arXiv preprint arXiv:2405.18492}, 2024.

\bibitem[Nasr et~al.(2023)Nasr, Carlini, Hayase, Jagielski, Cooper, Ippolito, Choquette-Choo, Wallace, Tram{\`e}r, and Lee]{nasr2023scalable}
Nasr, M., Carlini, N., Hayase, J., Jagielski, M., Cooper, A.~F., Ippolito, D., Choquette-Choo, C.~A., Wallace, E., Tram{\`e}r, F., and Lee, K.
\newblock Scalable extraction of training data from (production) language models.
\newblock \emph{arXiv preprint arXiv:2311.17035}, 2023.

\bibitem[Radford et~al.(2019)Radford, Wu, Child, Luan, Amodei, Sutskever, Dean, and Ghemawat]{radford2019language}
Radford, A., Wu, J., Child, R., Luan, D., Amodei, D., Sutskever, I., Dean, J., and Ghemawat, S.
\newblock Language models are unsupervised multitask learners.
\newblock In \emph{OSDI'04: Sixth Symposium on Operating System Design and Implementation}, pp.\  137--150, 2019.

\bibitem[Rahman \& Santacana(2023)Rahman and Santacana]{rahman2023beyond}
Rahman, N. and Santacana, E.
\newblock Beyond fair use: Legal risk evaluation for training {LLMs} on copyrighted text.
\newblock In \emph{ICML Workshop on Generative AI and Law}, 2023.

\bibitem[Team et~al.(2023)Team, Anil, Borgeaud, Wu, Alayrac, Yu, Soricut, Schalkwyk, Dai, Hauth, et~al.]{team2023gemini}
Team, G., Anil, R., Borgeaud, S., Wu, Y., Alayrac, J.-B., Yu, J., Soricut, R., Schalkwyk, J., Dai, A.~M., Hauth, A., et~al.
\newblock Gemini: a family of highly capable multimodal models.
\newblock \emph{arXiv preprint arXiv:2312.11805}, 2023.

\bibitem[Tirumala et~al.(2022)Tirumala, Markosyan, Zettlemoyer, and Aghajanyan]{tirumala2022memorization}
Tirumala, K., Markosyan, A., Zettlemoyer, L., and Aghajanyan, A.
\newblock Memorization without overfitting: Analyzing the training dynamics of large language models.
\newblock \emph{Advances in Neural Information Processing Systems}, 35:\penalty0 38274--38290, 2022.

\bibitem[Touvron et~al.(2023)Touvron, Martin, Stone, Albert, Almahairi, Babaei, Bashlykov, Batra, Bhargava, Bhosale, et~al.]{touvron2023llama}
Touvron, H., Martin, L., Stone, K., Albert, P., Almahairi, A., Babaei, Y., Bashlykov, N., Batra, S., Bhargava, P., Bhosale, S., et~al.
\newblock \text{LLaMa} 2: Open foundation and fine-tuned chat models.
\newblock \emph{arXiv preprint arXiv:2307.09288}, 2023.

\bibitem[Vyas et~al.(2023)Vyas, Kakade, and Barak]{vyas2023provable}
Vyas, N., Kakade, S.~M., and Barak, B.
\newblock On provable copyright protection for generative models.
\newblock In \emph{International Conference on Machine Learning}, pp.\  35277--35299. PMLR, 2023.

\bibitem[Wang et~al.(2023)Wang, Polo, Sun, Kundu, Xing, and Yurochkin]{wang2023fusing}
Wang, H., Polo, F.~M., Sun, Y., Kundu, S., Xing, E., and Yurochkin, M.
\newblock Fusing models with complementary expertise.
\newblock In \emph{Annual Conference on Neural Information Processing Systems}, 2023.

\bibitem[Wei et~al.(2024)Wei, Shi, Huang, Smith, Zhang, Zettlemoyer, Li, and Henderson]{wei2024evaluating}
Wei, B., Shi, W., Huang, Y., Smith, N.~A., Zhang, C., Zettlemoyer, L., Li, K., and Henderson, P.
\newblock Evaluating copyright takedown methods for language models.
\newblock \emph{arXiv preprint arXiv:2406.18664}, 2024.

\bibitem[Wortsman et~al.(2022)Wortsman, Ilharco, Gadre, Roelofs, Gontijo-Lopes, Morcos, Namkoong, Farhadi, Carmon, Kornblith, et~al.]{wortsman2022model}
Wortsman, M., Ilharco, G., Gadre, S.~Y., Roelofs, R., Gontijo-Lopes, R., Morcos, A.~S., Namkoong, H., Farhadi, A., Carmon, Y., Kornblith, S., et~al.
\newblock Model soups: averaging weights of multiple fine-tuned models improves accuracy without increasing inference time.
\newblock In \emph{International conference on machine learning}, pp.\  23965--23998. PMLR, 2022.

\bibitem[Yan et~al.(2022)Yan, Li, Guo, Li, Li, and Lin]{yan2022arcane}
Yan, H., Li, X., Guo, Z., Li, H., Li, F., and Lin, X.
\newblock {ARCANE}: An efficient architecture for exact machine unlearning.
\newblock In \emph{IJCAI}, volume~6, pp.\ ~19, 2022.

\bibitem[Yu et~al.(2023)Yu, Wu, Zhang, Wang, Vorobeychik, and Xiao]{yu2023codeiipprompt}
Yu, Z., Wu, Y., Zhang, N., Wang, C., Vorobeychik, Y., and Xiao, C.
\newblock {C}ode{IPP}rompt: Intellectual property infringement assessment of code language models.
\newblock In Krause, A., Brunskill, E., Cho, K., Engelhardt, B., Sabato, S., and Scarlett, J. (eds.), \emph{Proceedings of the 40th International Conference on Machine Learning}, volume 202 of \emph{Proceedings of Machine Learning Research}, pp.\  40373--40389. PMLR, 23--29 Jul 2023.
\newblock URL \url{https://proceedings.mlr.press/v202/yu23g.html}.

\bibitem[Zhang et~al.(2023)Zhang, Ippolito, Lee, Jagielski, Tram{\`e}r, and Carlini]{zhang2023counterfactual}
Zhang, C., Ippolito, D., Lee, K., Jagielski, M., Tram{\`e}r, F., and Carlini, N.
\newblock Counterfactual memorization in neural language models.
\newblock \emph{Advances in Neural Information Processing Systems}, 36:\penalty0 39321--39362, 2023.

\bibitem[Zhang et~al.(2024{\natexlab{a}})Zhang, Lin, Bai, and Mei]{zhang2024negative}
Zhang, R., Lin, L., Bai, Y., and Mei, S.
\newblock Negative preference optimization: From catastrophic collapse to effective unlearning.
\newblock \emph{arXiv preprint arXiv:2404.05868}, 2024{\natexlab{a}}.

\bibitem[Zhang et~al.(2024{\natexlab{b}})Zhang, Luo, Yuan, and Yao]{zhang2024autonomous}
Zhang, Y., Luo, Y., Yuan, Y., and Yao, A.~C.
\newblock Autonomous data selection with language models for mathematical texts.
\newblock In \emph{ICLR 2024 Workshop on Navigating and Addressing Data Problems for Foundation Models}, 2024{\natexlab{b}}.

\end{thebibliography}
\bibliographystyle{icml2024}

\clearpage
\appendix
\onecolumn
\section*{Appendices}
The following appendices provide additional results and discussions, deferred proofs, and experimental details.
\FloatBarrier
\section{Additional experiments}

\subsection{Additional metrics for copyright infringement}
\label{app:more_copyright}

We report the copyright infringement results for the StarCoder and LLaMa2 datasets, along with additional metrics. Specifically, we include Jaccard and cosine similarities and the METEOR score to measure approximate memorization, as well as semantic similarity for a more high-level measure that does not necessarily indicate copyright infringement. 

Results using Jaccard and cosine similarities and the METEOR score confirm the observations from the main text, closely aligning with previous metrics. The semantic similarity for CP-Fuse and $\cpdelta$ remains consistently high, comparable to that of the overfitted models, suggesting that no semantic information is lost when applying these methods.

\begin{table}[ht]
\centering
\caption{Copyright-infringement metrics averaged at the 95th percentile for StarCoder and LLaMa2 across different data splits. The table presents results for the overfitted models, $\cpdelta$, and CP-Fuse. Metrics include Exact Matching (EM), Normalized Levenshtein Distance (LEV), Jaccard Similarity (JAC), Cosine Similarity (COS), Semantic Similarity (SEM), ROUGE-L (ROU), BLEU Score (BLE), METEOR Score (MET), and Infringement Count (IC\textsubscript{50}, IC\textsubscript{160}). $\downarrow$ Means lower is better, $\uparrow$ means higher is better.}

\begin{tabular}{ll|cccccccccc}
\toprule
& & \textbf{EM} $\downarrow$ & \textbf{IC\textsubscript{50}} $\downarrow$ & \textbf{IC\textsubscript{160}} $\downarrow$ & \textbf{ROU} $\downarrow$ & \textbf{BLE} $\downarrow$ & \textbf{MET} $\downarrow$ & \textbf{JAC} $\downarrow$ & \textbf{COS} $\downarrow$ & \textbf{SEM} $\downarrow$ & \textbf{LEV} $\uparrow$ \\
\midrule
\multicolumn{2}{c|}{\textbf{StarCoder}} \\
\midrule
\multirow{3}{*}{\makecell{Overfit \\ Split 1}}
& Split 1 & 2489.28 & 2439.96 & 2329.96 & 1.00 & 1.00 & 1.00 & 1.00 & 1.00 & 1.00 & 0.00 \\
& Split 2 & 33.74 & 1.49 & 0.12 & 0.54 & 0.77 & 0.46 & 0.36 & 0.66 & 0.96 & 0.54 \\
& Test & 65.88 & 1.65 & 0.34 & 0.55 & 0.80 & 0.45 & 0.35 & 0.63 & 0.96 & 0.49 \\
\midrule
\multirow{3}{*}{\makecell{Overfit \\ Split 2}}
& Split 1 & 47.88 & 1.47 & 0.31 & 0.53 & 0.82 & 0.47 & 0.41 & 0.70 & 0.96 & 0.52 \\
& Split 2 & 2182.16 & 2129.48 & 2019.48 & 1.00 & 1.00 & 1.00 & 1.00 & 1.00 & 1.00 & 0.00 \\
& Test & 41.38 & 1.24 & 0.17 & 0.53 & 0.66 & 0.38 & 0.30 & 0.62 & 0.95 & 0.52 \\
\midrule
\midrule
\multirow{3}{*}{CP-Fuse}
& Split 1 & 59.48 & 66.89 & 0.602 & 0.81 & 0.66 & 0.67 & 0.52 & 0.74 & 0.99 & 0.23 \\
& Split 2 & 48.88 & 101.16 & 1.304 & 0.81 & 0.64 & 0.69 & 0.56 & 0.80 & 0.99 & 0.24 \\
& Test & 35.59 & 12.25 & 0.04 & 0.57 & 0.71 & 0.36 & 0.27 & 0.59 & 0.95 & 0.52 \\
\midrule
\multirow{3}{*}{CP-$\Delta$}
& Split 1 & 341.60 & 312.28 & 136.52 & 1.00 & 1.00 & 1.00 & 1.00 & 1.00 & 1.00 & 0.07 \\
& Split 2 & 162.80 & 337.28 & 152.64 & 1.00 & 1.00 & 1.00 & 1.00 & 1.00 & 1.00 & 0.02 \\
& Test & 39.91 & 1.20 & 0.03 & 0.58 & 0.80 & 0.40 & 0.31 & 0.61 & 0.96 & 0.51 \\
\midrule
\multicolumn{2}{c|}{\textbf{LLaMa2}} \\
\midrule
\multirow{3}{*}{\makecell{Overfit \\ Split 1}}
& Split 1 & 1397.68 & 1595.12 & 1482.84 & 1.00 & 1.00 & 1.00 & 1.00 & 1.00 & 1.00 & 0.00 \\
& Split 2 & 31.00 & 0.016 & 0.00 & 0.25 & 0.06 & 0.24 & 0.17 & 0.76 & 0.98 & 0.70 \\
& Test & 28.76 & 0.00 & 0.00 & 0.22 & 0.15 & 0.23 & 0.16 & 0.75 & 0.98 & 0.70 \\
\midrule
\multirow{3}{*}{\makecell{Overfit \\ Split 2}}
& Split 1 & 42.12 & 0.17 & 0.00 & 0.27 & 0.08 & 0.27 & 0.19 & 0.77 & 0.98 & 0.68 \\
& Split 2 & 1570.88 & 1798.72 & 1688.72 & 1.00 & 1.00 & 1.00 & 1.00 & 1.00 & 1.00 & 0.00 \\
& Test & 37.48 & 0.068 & 0.00 & 0.26 & 0.07 & 0.26 & 0.19 & 0.76 & 0.98 & 0.69 \\
\midrule
\midrule
\multirow{3}{*}{CP-Fuse}
& Split 1 & 94.20 & 15.14 & 0.00 & 0.35 & 0.14 & 0.30 & 0.26 & 0.80 & 0.98 & 0.62 \\
& Split 2 & 65.84 & 0.378 & 0.00 & 0.34 & 0.14 & 0.31 & 0.24 & 0.81 & 0.98 & 0.63 \\
& Test & 47.92 & 0.008 & 0.00 & 0.28 & 0.07 & 0.25 & 0.19 & 0.77 & 0.98 & 0.68 \\
\midrule
\multirow{3}{*}{CP-$\Delta$}
& Split 1 & 273.20 & 408.72 & 253.4 & 0.72 & 0.58 & 0.64 & 0.61 & 0.89 & 0.99 & 0.30 \\
& Split 2 & 284.80 & 162.6 & 1.658 & 0.50 & 0.33 & 0.40 & 0.35 & 0.86 & 0.98 & 0.51 \\
& Test & 57.50 & 0.088 & 0.00 & 0.29 & 0.07 & 0.26 & 0.21 & 0.77 & 0.98 & 0.67 \\
\bottomrule
\end{tabular}
\label{tab:all_copyright}
\end{table}

\newpage

\subsection{Experiments with GPT-2 XL and Phi-2}
\label{app:gpt_results}
We present additional results with GPT-2 XL, a 1.5B parameter version of GPT-2, and the Phi-2 model \citep{javaheripi2023phi}. These models are smaller than the ones discussed in the main text, and thus, we expect that they exhibit lower memorization rates \citep{tirumala2022memorization}. We report exact matching for copyright infringement and the perplexity score to measure utility. 

\Cref{tab:combined_results_gpt} shows a similar trend compared to the results from \Cref{sec:experiments}. Specifically, the $\cpdelta$ baseline demonstrates memorization of strings that are twice as large as those produced by our method. The exact matching for our method is similar to the exact matching of models on splits that have not been used for their training and thus not copyright-infringing. Furthermore, both our method and $\cpdelta$ show competitive perplexity.

\begin{table*}[ht!]
\centering
\caption{Perplexity (PPL) and Exact Matching (EM) at the 95th percentile for GPT-2 XL and Phi-2 across fine-tuning and test splits. We report results for the overfitted models, $\cpdelta$, and CP-Fuse.}
\begin{tabular}{ll|cc|cc}
\toprule
& & \multicolumn{2}{c|}{\textbf{GPT-2 XL}} & \multicolumn{2}{c}{\textbf{Phi-2}} \\
\textbf{Model} & \textbf{Split} & \textbf{PPL} & \textbf{EM} & \textbf{PPL} & \textbf{EM} \\
\midrule
\multirow{3}{*}{\makecell{Overfit \\ Split 1}} 
& Split 1 & 1.10 & 1521.76 & 1.24 & 1369.16 \\
& Split 2 & 1.44 & 38.48 & 1.34 & 33.55 \\
& Test & 1.44 & 39.80 & 1.35 & 30.04 \\
\midrule
\multirow{3}{*}{\makecell{Overfit \\ Split 2}} 
& Split 1 & 1.45 & 37.14 & 1.33 & 29.80 \\
& Split 2 & 1.28 & 1344.20 & 1.23 & 1296.04 \\
& Test & 1.45 & 39.18 & 1.33 & 32.27 \\
\midrule
\midrule
\multirow{3}{*}{CP-Fuse} 
& Split 1 & 1.51 & 45.24 & 1.46 & 41.76 \\
& Split 2 & 1.51 & 57.61 & 1.46 & 45.96 \\
& Test & 1.51 & 40.48 & 1.49 & 34.50 \\
\midrule
\multirow{3}{*}{CP-$\Delta$} 
& Split 1 & 1.48 & 72.54 & 1.41 & 82.44 \\
& Split 2 & 1.47 & 113.20 & 1.41 & 89.12 \\
& Test & 1.49 & 42.79 & 1.44 & 36.18 \\
\bottomrule
\end{tabular}
\label{tab:combined_results_gpt}
\end{table*}

\subsection{Extended experimental results from \Cref{tab:results_utility}}
\label{app:complete_results}

For completeness, we include additional results on the different datasets in \Cref{tab:results_perplexity} and \ref{tab:results_apps}. In \Cref{tab:results_perplexity}, we present the perplexity scores obtained with StarCoder on the Python instructions dataset and LLaMa2 on the Math abstracts dataset across all splits. The results are consistent with previous observations, demonstrating that CP-Fuse maintains competitive perplexity scores across different splits.

\begin{table*}
\centering
\begin{minipage}{.45\linewidth}
  \centering
  \begin{tabular}{ll|cc}
    \toprule
    & & \multicolumn{2}{c}{\textbf{Perplexity}} \\
    \midrule
    \textbf{Model} & \textbf{Split} & \makecell{\textbf{Python} \\ \textbf{Instructions}} & \makecell{\textbf{Math} \\ \textbf{Abstracts}} \\
    \midrule
    \multirow{3}{*}{\makecell{Overfit \\ Split 1}} 
    & Split 1 & 1.01 & 1.22 \\
    & Split 2 & 1.13 & 1.43 \\
    & Test & 1.12 & 1.41 \\
    \midrule
    \multirow{3}{*}{\makecell{Overfit \\ Split 2}} 
    & Split 1 & 1.13 & 1.23 \\
    & Split 2 & 1.01 & 1.01 \\
    & Test & 1.13 & 1.23 \\
    \midrule
    \midrule
    \multirow{3}{*}{CP-Fuse} & Split 1 & 1.17 & 1.59 \\
    & Split 2 & 1.17 & 1.61 \\
    & Test & 1.18 & 1.61 \\
    \midrule
    \multirow{3}{*}{$\cpdelta$} & Split 1 & 1.12 & 1.45 \\
    & Split 2 & 1.11 & 1.47 \\
    & Test & 1.16 & 1.54 \\
    \bottomrule
  \end{tabular}
  \caption{Perplexity metrics for different methods on Python instructions and Math abstracts datasets across various splits.}
  \label{tab:results_perplexity}
\end{minipage}%
\hfill
\begin{minipage}{.45\linewidth}
  \centering
  \begin{tabular}{ll|cc}
    \toprule
    & & \multicolumn{2}{c}{\textbf{APPS Dataset}} \\
    \midrule
    \textbf{Model} & \textbf{Split} & \makecell{\textbf{Perplexity}} & \makecell{\textbf{Exact} \\ \textbf{Matching}} \\
    \midrule
    \multirow{3}{*}{\makecell{Overfit \\ Split 1}} 
    & Split 1 &  1.11 &  333.64 \\
    & Split 2 &  1.15 &  113.56 \\
    & Test &   1.16 &  57.91 \\
    \midrule
    \multirow{3}{*}{\makecell{Overfit \\ Split 2}} 
    & Split 1 &  1.15 &  104.36 \\
    & Split 2 &  1.11 &  322.64\\
    & Test &  1.19 &  58.00  \\
    \midrule
    \midrule
    \multirow{3}{*}{CP-Fuse}  & Split 1 & 1.14 &  104.67 \\
    & Split 2 & 1.14 &  113.88\\
    & Test & 1.16 & 57.18 \\
    \midrule
    \multirow{3}{*}{$\cpdelta$} & Split 1 & 1.14 & 137.00 \\
    & Split 2 & 1.14 &  140.04\\
    & Test &  1.17 &  58.50 \\
    \bottomrule
  \end{tabular}
  \caption{Perplexity and Exact Matching metrics for different methods on the APPS dataset across various splits.}
  \label{tab:results_apps}
\end{minipage}
\end{table*}

\begin{table*}
\centering
\caption{Perplexity (PPL) and Exact Matching (EM) at the 95th percentile for StarCoder, Phi-2, GPT-2 XL, and LLaMa2 across fine-tuning and test splits. We report results for the early-stopped (ES) models, the baseline $\cpdelta$, and our method CP-Fuse.}
\begin{tabular}{ll|cc|cc|cc|cc}
\toprule
& & \multicolumn{2}{c|}{\textbf{StarCoder}} & \multicolumn{2}{c|}{\textbf{Phi-2}} & \multicolumn{2}{c|}{\textbf{GPT-2 XL}} & \multicolumn{2}{c}{\textbf{LLaMa2}} \\
\textbf{Model} & \textbf{Split} & \textbf{PPL} & \textbf{EM} & \textbf{PPL} & \textbf{EM} & \textbf{PPL} & \textbf{EM} & \textbf{PPL} & \textbf{EM} \\
\midrule
\multirow{3}{*}{\makecell{ES Split 1}} & Split 1 & 1.26 & 159.36 & 1.56 & 41.71 & 1.79 & 65.83 & 1.46 & 207.44 \\
& Split 2 & 1.30 & 39.23 & 1.60 & 41.08 & 1.78 & 41.68 & 1.50 & 46.87 \\
& Test & 1.30 & 51.71 & 1.60 & 42.35 & 1.82 & 39.68 & 1.52 & 44.83 \\
\midrule
\multirow{3}{*}{\makecell{ES Split 2}} & Split 1 & 1.25 & 31.96 & 1.66 & 45.71 & 1.60 & 38.60 & 1.49 & 44.76 \\
& Split 2 & 1.24 & 145.04 & 1.67 & 46.56 & 1.59 & 60.60 & 1.40 & 280.20 \\
& Test & 1.27 & 43.74 & 1.67 & 40.88 & 1.60 & 40.78 & 1.47 & 44.65 \\
\midrule
\midrule
\multirow{3}{*}{CP-Fuse} & Split 1 & 1.29 & 46.96 & 1.58 & 44.10 & 1.69 & 43.82 & 1.52 & 52.21 \\
& Split 2 & 1.29 & 44.50 & 1.61 & 43.58 & 1.71 & 51.62 & 1.52 & 53.30 \\
& Test & 1.29 & 49.43 & 1.59 & 41.62 & 1.73 & 43.78 & 1.53 & 45.00 \\
\midrule
\multirow{3}{*}{CP-$\Delta$} & Split 1 & 1.29 & 70.17 & 1.50 & 44.77 & 1.70 & 50.14 & 1.46 & 68.84 \\
& Split 2 & 1.29 & 59.04 & 1.54 & 46.96 & 1.70 & 49.00 & 1.45 & 61.48 \\
& Test & 1.30 & 48.12 & 1.55 & 42.38 & 1.70 & 43.00 & 1.46 & 45.79 \\
\bottomrule
\end{tabular}
\label{tab:combined_results_mild}
\end{table*}

\Cref{tab:results_apps} provides further analysis by showing the perplexity scores and exact matching rates above the 95th percentile for StarCoder models trained on the APPS dataset, evaluated on both the APPS fine-tuning and test sets. CP-Fuse continues to be effective in reducing regurgitation while maintaining low perplexity, comparable to that of the overfitted models.

\subsection{Additional experiments with early-stopped models}
\label{app:early_stopped}

In this section, we present the additional experimental results with early-stopped models, including the GPT-2 XL and Phi-2 models. Specifically, we stop fine-tuning upon detecting an increase in memorization, as is a common practice in the literature \citep{mireshghallah2022memorization}. \Cref{tab:combined_results_mild}  shows that the early-stopped models exhibit higher perplexity (i.e., worse) compared to CP-Fuse applied to heavily overfitted models (refer to the main results in \Cref{tab:results_copyright}). Moreover, early-stopped models for Phi-2 and GPT-2 XL show similar exact memorization at the 95th percentile than CP-Fuse in the text-based task.

Additionally, we apply both the baseline $\cpdelta$ and CP-Fuse on top of the early-stopped models. We observe that CP-Fuse further reduces regurgitation of memorized training samples (e.g., StarCoder by a factor of 3) and, in some cases, improves perplexity (e.g., Phi-2), while consistently outperforming $\cpdelta$.

\subsection{Ablation studies for the grid size}

We conduct ablation studies on the grid size used for solving the optimization problem in \Cref{eq:opt}. Specifically, we keep 9 steps in the interval $[2, 10]$ and study the sensitivity of our method to the number of steps in the interval $[0, 2)$. 

\Cref{tab:ablation_grid} shows the perplexity and average exact matching (above the 95th and 99th percentiles) for different numbers of steps. Remarkably, for StarCoder and Phi-2, we observe similar levels of memorization while perplexity decreases (i.e., better) for smaller grids. Note that using smaller grids significantly accelerates the decoding process. Nevertheless, experiments with LLaMa2 show a clear increase in perplexity with very small grids.

\begin{table*}[h]
\centering
\caption{Ablation Study: Perplexity (PPL) and Exact Matching (EM) at the 95th and 99th percentiles for StarCoder, Phi-2, and LLaMa2 with different grid sizes.}
\begin{tabular}{ll|ccc|ccc|ccc}
\toprule
& & \multicolumn{3}{c|}{\textbf{StarCoder}} & \multicolumn{3}{c|}{\textbf{Phi-2}} & \multicolumn{3}{c}{\textbf{LLaMa2}} \\
\textbf{Grid Size} & \textbf{Split} & \textbf{PPL} & \textbf{EM\textsubscript{95}} & \textbf{EM\textsubscript{99}} & \textbf{PPL} & \textbf{EM\textsubscript{95}} & \textbf{EM\textsubscript{99}} & \textbf{PPL} & \textbf{EM\textsubscript{95}} & \textbf{EM\textsubscript{99}} \\
\midrule
\multirow{3}{*}{2 + 9}
& Split 1 & 1.09 & 86.75 & 180.40 & 1.18 & 45.56 & 54.80 & 2.41 & 57.52 & 70.60 \\
& Split 2 & 1.09 & 81.08 & 146.40 & 1.18 & 44.39 & 54.60 & 2.52 & 46.04 & 56.75 \\
& Test & 1.10 & 47.42 & 87.20 & 1.19 & 34.65 & 42.20 & 2.52 & 36.84 & 47.60 \\
\midrule
\multirow{3}{*}{5 + 9}
& Split 1 & 1.17 & 94.20 & 226.20 & 1.39 & 45.30 & 55.50 & 1.59 & 59.48 & 74.80 \\
& Split 2 & 1.17 & 65.84 & 100.80 & 1.40 & 45.90 & 57.40 & 1.61 & 48.88 & 60.40 \\
& Test & 1.18 & 47.92 & 88.80 & 1.40 & 33.84 & 45.33 & 1.64 & 34.95 & 40.25 \\
\midrule
\multirow{3}{*}{10 + 9}
& Split 1 & 1.19 & 89.88 & 201.00 & 1.46 & 41.76 & 52.00 & 1.63 & 55.54 & 65.40 \\
& Split 2 & 1.19 & 72.92 & 132.80 & 1.46 & 45.96 & 55.40 & 1.64 & 48.74 & 55.60 \\
& Test & 1.20 & 48.80 & 93.20 & 1.49 & 34.50 & 42.20 & 1.67 & 35.59 & 42.00 \\
\midrule
\multirow{3}{*}{20 + 9}
& Split 1 & 1.20 & 90.42 & 201.80 & 1.51 & 44.82 & 59.80 & 1.65 & 57.67 & 70.50 \\
& Split 2 & 1.21 & 70.48 & 115.00 & 1.51 & 46.57 & 56.60 & 1.68 & 48.45 & 56.80 \\
& Test & 1.21 & 47.64 & 91.80 & 1.54 & 35.29 & 43.80 & 1.68 & 35.21 & 44.00 \\
\bottomrule
\end{tabular}
\label{tab:ablation_grid}
\end{table*}

\subsection{Visualizing the balancing property and the adaptively selected parameters $\alpha_t$ and $\beta_t$}
\label{app:weights}

\begin{figure}[ht]
    \centering
        \includegraphics[width=0.6\textwidth]{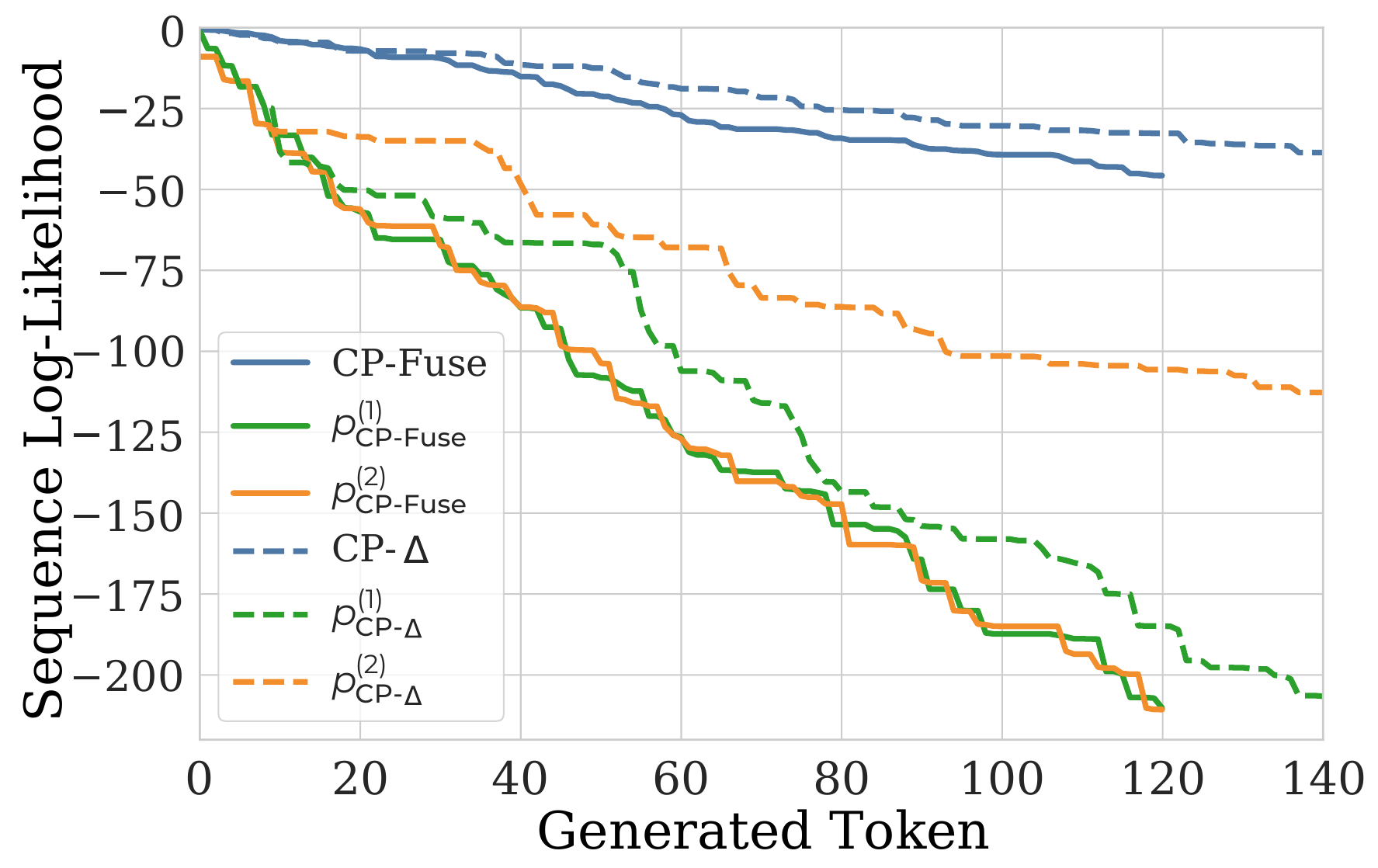}
\caption{(Same as \Cref{fig:cumulative_main}) Log-likelihood for the sequence produced by CP-Fuse and $\cpdelta$, and the corresponding base models $p^{(1)}$ and $p^{(2)}$ at each token in greedy decoding. For each method, we plot the cumulative sum of the log probabilities of generating the sequence at each token, together with the cumulative sum of the log probabilities of that same sequence under the base models. Due to the balancing property, CP-Fuse achieves $\log p^{(1)}(y_{\leq t} \vert x) \approx \log p^{(2)}(y_{\leq t} \vert x)$ at all steps of the generation, indicating that the tokens produced by CP-Fuse are roughly equally likely under both base models, hence preventing the reproduction of memorized samples. In contrast, $\cpdelta$ places significantly more weight on the second model $p^{(2)}$, as evidenced by the much higher log-likelihood of the generated tokens under $p^{(2)}$ compared to $ p^{(1)}$. This increases the likelihood of reproducing memorized samples from $p^{(2)}$.}
    \label{fig:cumulative}
\end{figure}

In \Cref{fig:cumulative}, we plot the log densities $\log p(y_{\leq t} \vert x)$, $\log p^{(1)}(y_{\leq t} \vert x)$, and $\log p^{(2)}(y_{\leq t} \vert x)$ for both CP-Fuse and CP-$\Delta$ for a sequence generated by both models given a prompt $x$ contained in the second fine-tuning data split. As we can see, for CP-Fuse, the balancing property from Lemma~\ref{lem:balance} ensures that the generated sequence has approximately the same log probability for both base models, $\log p^{(1)}(y_{\leq t} \vert x) \approx \log p^{(2)}(y_{\leq t} \vert x)$. In contrast, the sequence generated by $\cpdelta$ occurs more likely under $\log p^{(2)}(y_{\leq t} \vert x)$, which overfitted on the prompt $x$, than $\log p^{(1)}(y_{\leq t} \vert x)$. This makes $\cpdelta$ more vulnerable to replicating text memorized by $\log p^{(2)}(y_{\leq t} \vert x)$, as we observed in our experimental results.

\begin{figure}[b!]
    \centering
    \begin{minipage}{0.48\textwidth}
        \centering
        \includegraphics[width=\textwidth]{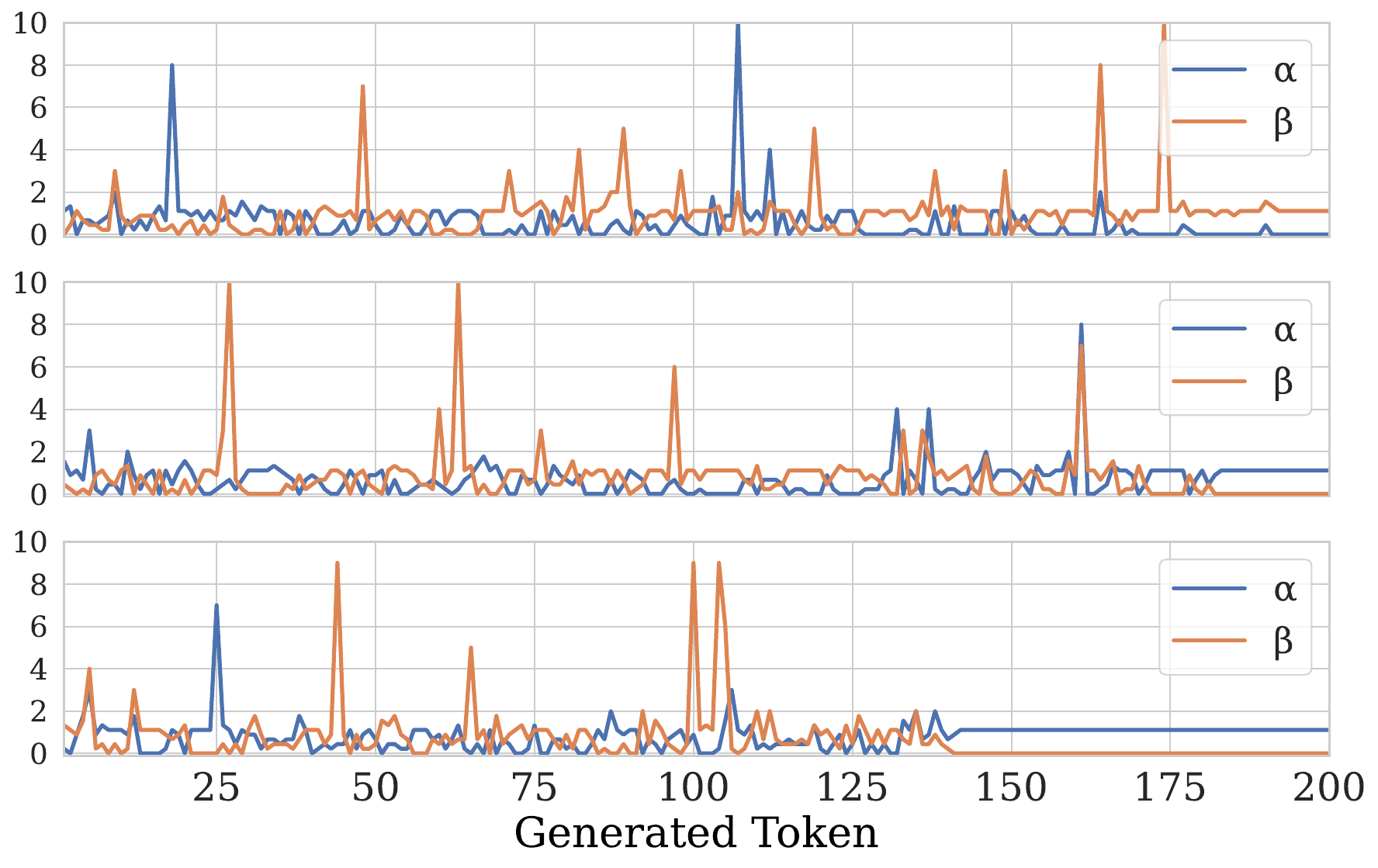}
        \label{fig:}
    \end{minipage}
    \begin{minipage}{0.48\textwidth}
        \centering
        \includegraphics[width=\textwidth]{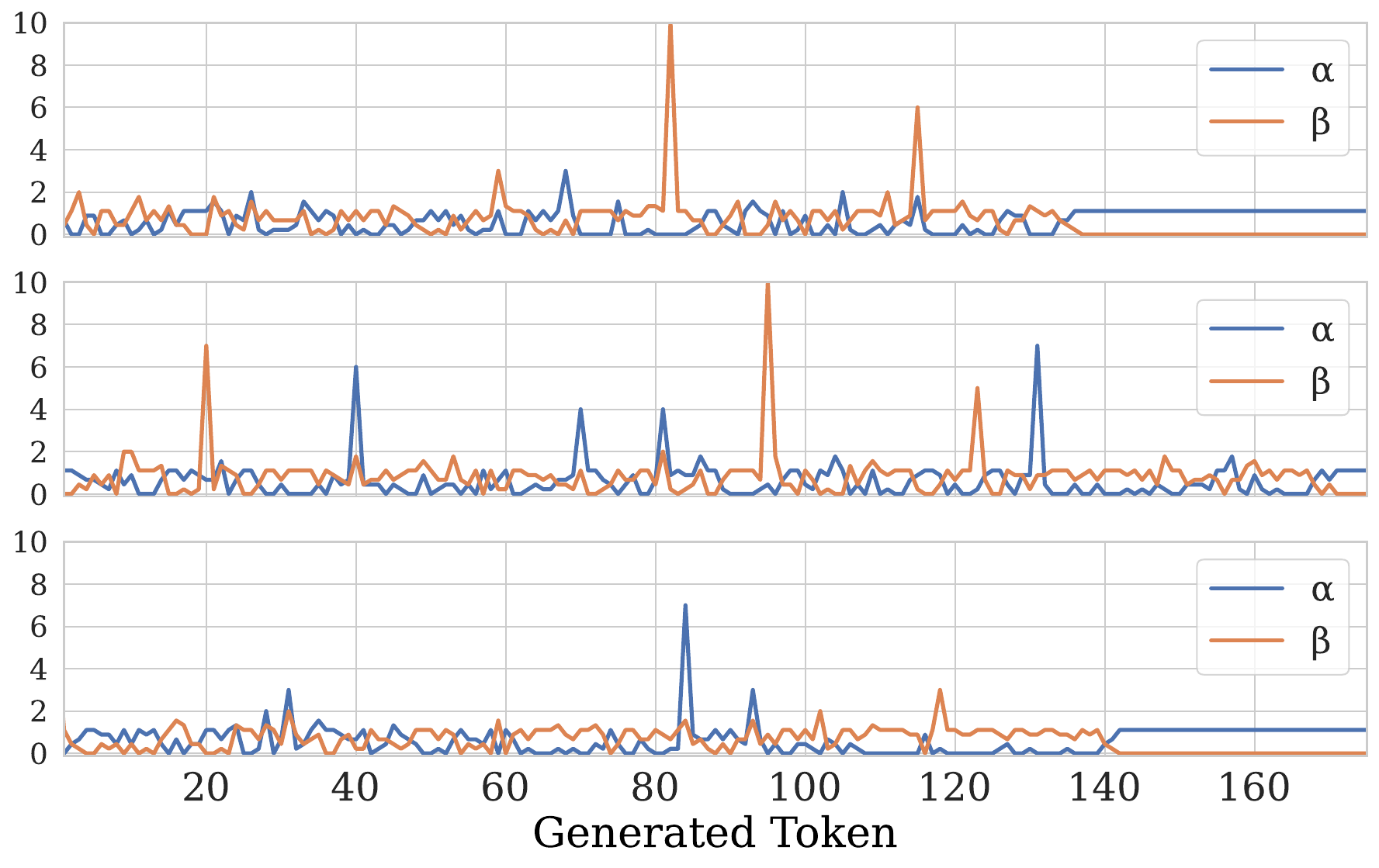}
        \label{fig:figure2}
    \end{minipage}
    \caption{Evolution of the parameters $\alpha_t$ and $\beta_t$ during greedy decoding. We randomly sampled six examples of text generated by our method CP-Fuse, combining overfitted Phi-2 models on the math abstract dataset. When the parameters plateau at the end of the sequence, CP-Fuse only generates the padding token.}
    \label{fig:weights}
\end{figure}

In \Cref{fig:weights}, we illustrate how the parameters $\alpha_t$ and $\beta_t$ adaptively change during the generation of an output via greedy decoding. We observe the consequences of the balancing property (\Cref{lem:balance}): when one model heavily dominates the generation process, our algorithm increases the weight of the other model to the point that the generation is independent of the dominating model. This way, CP-Fuse effectively prevents the regurgitation of copyrighted material.

\subsection{Examples of outputs generated by CP-Fuse}

In this section, we present output examples generated by our method and compare them with outputs from the copyright-infringing overfitted model, the baseline $\cpdelta$, the early-stopped model, and the base model without fine-tuning. All examples are randomly sampled from the fine-tuning datasets.

Figures \ref{fig:example1_code_methods}, \ref{fig:example2_code_methods}, and \ref{fig:example3_code_methods} show outputs generated for the Python instructional dataset. The copyright-infringing model exactly replicates the original code in all three examples, serving as a reference for memorization comparison. The $\cpdelta$ algorithm produces code closely resembling the original, with a nearly exact full match in \Cref{fig:example1_code_methods} and \ref{fig:example3_code_methods}, and an exact reproduction of a comment with a link in \Cref{fig:example2_code_methods}. In contrast, CP-Fuse generates significantly different code that is correct and arguably of higher quality, incorporating exception handling and new features, such as the selection of different statistics in \Cref{fig:example3_code_methods}. The early-stopped model produces low-quality code, often oversimplifying tasks (\Cref{fig:example1_code_methods} and \ref{fig:example3_code_methods}) and committing syntax errors, such as an open quotation in \Cref{fig:example2_code_methods}. Finally, the base model often fails to generate code and produces natural or nonsensical text instead (\Cref{fig:example1_code_methods} and \ref{fig:example3_code_methods}), highlighting the necessity of fine-tuning in the first place, but also can generate a correct output as in  \Cref{fig:example2_code_methods}.

Figures \ref{fig:example1_text_methods}, \ref{fig:example2_text_methods} and \ref{fig:example3_text_methods} illustrate outputs generated for the task of producing abstracts from math paper titles. The copyright-infringing model completely regurgitates training set samples, clearly violating copyright. Both CP-Fuse and $\cpdelta$ generate reasonable, high-quality text; however, $\cpdelta$ reproduces substantially more memorized text compared to CP-Fuse. Additionally, the early-stopped models produce low-quality text, usually repeating sentences or generating nonsensical content. Finally, the base model fails to generate coherent text.

\begin{figure}[!htbp]
    \centering
    \begin{promptmdf}
        \textbf{Copyright-infringing model:}
    \begin{lstlisting}[style=mystyle, language=Python]
def set_data(self, data, addr=0):
    if self._mem_bytes < len(data):
        raise ValueError('Size of data (%d bytes) is too big for memory (%d bytes)' % (len(data), self._mem_bytes))
    self._intf.write(self._conf['base_addr'] + self._spi_mem_offset + addr, data)
        \end{lstlisting}
        \textbf{CP-Fuse:}
        \begin{lstlisting}[style=mystyle, language=Python]
def set_data(self, data):
    if self._data_bytes:
        self._buf.write(data)
        return
    try:
        self._buf = BytesIO(data)
    except TypeError:
        self._buf = StringIO(data)
        \end{lstlisting}
        \textbf{$\cpdelta$:}
        \begin{lstlisting}[style=mystyle, language=Python]
def set_data(self, data, addr=0):
    if self._mem_bytes < len(data):
        raise ValueError('Size of data (%d bytes) is too big for memory (%d bytes)' % (len(data), self._mem_bytes))
    self._intf.write(self._conf['base_addr'] + addr, data)
        \end{lstlisting}
        \textbf{Early-stopped model:}
        \begin{lstlisting}[style=mystyle, language=Python]
def set_data(self, data):
    self.data = data
    self.update()
        \end{lstlisting}
        \textbf{Base model:} \\
 <details><summary><b>Answer</b>
        </summary> <p>  Answer: B  ```python stream.data = data ```  </p> </details>  ---  Q10. Which of the following is a valid way to create a 
        dictionary in Python 3?  -   [ ] `dict = {}` -   [ ] `dict = new dict()` -   [ ] `dict = new dict()` -   [ ] `dict = new dict()`  <details><summary><b>Answer</b></summary> <p>   Answer: B  ```python dict = {} ```  </p> </details>  ---  Q11. [...]
    \end{promptmdf}

    \caption{Example of text generated by the overfitted, copyright-infringing model, CP-Fuse, $\cpdelta$, the early-stopped model, and the base model for the Python instructional dataset using StarCode models.}
    \label{fig:example1_code_methods}
\end{figure}

\begin{figure}[!htbp]
    \centering
    \begin{promptmdfthree}
        \textbf{Copyright-infringing model:}
    \begin{lstlisting}[style=mystyle, language=Python]
def calc_mean_hr(rr, fs=None, min_rr=None, max_rr=None, rr_units='samples'):
    if not len(rr):
        return 0
    if min_rr is not None:
        rr = rr[rr > min_rr]
    if max_rr is not None:
        rr = rr[rr < max_rr]
    mean_rr = np.mean(rr)
    mean_hr = 60 / mean_rr
    # Convert to bpm
    if rr_units == 'samples':
        mean_hr = mean_hr * fs
    return mean_hr
        \end{lstlisting}
        \textbf{CP-Fuse:}
        \begin{lstlisting}[style=mystyle, language=Python]
def calc_mean_hr(self, rr, fs=None, window=None, n_fft=None, mean_type='mean', **kwargs):
    if not self.is_rr_set(rr):
        return
    if fs is None:
        fs = self.fs
    if window is None:
        window = self.window
    if n_fft is None:
        n_fft = self.n_fft
    if mean_type =='mean':
        mean = np.mean(rr)
    elif mean_type =='median':
        mean = np.median(rr)
    elif mean_type =='mode':
        mean = np.mode(rr)
    else:
        raise ValueError(
            "Unknown mean_type: {}".format(mean_type))
    mean_hr = 60 / self.duration(fs=fs, window=window, n_fft=n_fft) * mean
    return mean_hr
        \end{lstlisting}
        \textbf{$\cpdelta$:}
        \begin{lstlisting}[style=mystyle, language=Python]
def calc_mean_hr(rr, fs=None, min_rr=None, max_rr=None, rr_units='samples'):
    if not len(rr):
        return 0
    if min_rr is not None:
        rr = rr[rr > min_rr]
    if max_rr is not None:
        rr = rr[rr < max_rr]
    mean_rr = np.mean(rr)
    mean_hr = 60 / mean_rr
    # Convert to bpm
    if rr_units =='samples':
        mean_hr = mean_hr * fs
    return mean_hr
        \end{lstlisting}
        \textbf{Early-stopped model:}
                \begin{lstlisting}[style=mystyle, language=Python]
def mean_rr_interval(rr_intervals):
    mean_rr_interval = 0
    if len(rr_intervals) > 0:
        mean_rr_interval = sum(rr_intervals) / len(rr_intervals)
    return mean_rr_interval
        \end{lstlisting}
        \textbf{Base model:}\\ 
            The file format is a simple text file with one number per line.  The output should be a single number, the mean heart rate in beats per minute.  The program should be able to handle a file with any number of lines.  The program should be able to handle a file with no lines.  The program should be able to handle a file with a single line.  The program should be able to handle a file with a single number. [...]
    \end{promptmdfthree}
    \caption{Example of text generated by the overfitted, copyright-infringing model, CP-Fuse, $\cpdelta$, the early-stopped model, and the base model for the Python instructional dataset using StarCode models.}
    \label{fig:example3_code_methods}
\end{figure}

\begin{figure}[!htbp]
    \centering
    \begin{promptmdftwo}
        \textbf{Copyright-infringing model:}
    \begin{lstlisting}[style=mystyle, language=Python]
def read_file(filename):
    # intentionally *not* adding an encoding option to open
    # see here:
    # https://github.com/pypa/virtualenv/issues/201#issuecomment-3145690
    here = os.path.abspath(os.path.dirname(__file__))
    with codecs.open(os.path.join(here, 'graphql_compiler', filename), 'r') as f:
        return f.read()
        \end{lstlisting}
        \textbf{CP-Fuse:}
        \begin{lstlisting}[style=mystyle, language=Python]
def read_file(path):
    # intentionally ignoring errors here, as we want to know if the file exists
    package_file = open(path, "r")
    package_content = package_file.read()
    package_file.close()
    match = re.search(r"^name:.* version:.*", package_content)
    if match:
        name = match.group(1)
        version = match.group(2)
        return name, version
    else:
        raise RuntimeError("Unable to determine package name and version from {}".format(path))
        \end{lstlisting}
        \textbf{$\cpdelta$:}
        \begin{lstlisting}[style=mystyle, language=Python]
def read_file(filename):
    # intentionally *not* adding an encoding option to open
    # see here:
    # https://github.com/pypa/virtualenv/issues/201#issuecomment-3145690
    here = os.path.abspath(os.path.dirname(__file__))
    with open(os.path.join(here, 'VERSION')) as f:
        version = f.read().strip()
    return version
        \end{lstlisting}
        \textbf{Early-stopped model:}
        \begin{lstlisting}[style=mystyle, language=Python]
def read_package_file(filename):
    with open(filename, 'r') as f:
        content = f.read()
    match = re.search(r'\s*name\s*=\s*([\'"])(.*?)\1, content)
    if match is not None:
        name = match.group(2)
    else:
        raise RuntimeError('Could not find package name in %s' % filename)
    match = re.search(r'\s*version\s*=\s*([\'"])(.*?)\1', content)
    if match is not None:
        version = match.group(2)
    else:
        raise RuntimeError('Could not find package version in %s' % filename)
    return name, version
        \end{lstlisting}
        \textbf{Base model:}
        \begin{lstlisting}[style=mystyle, language=Python]
def read_pkg_file(filename):    
    with open(filename, 'r') as f:
        return f.read()
def get_name_and_version(filename):
    pkg_file = read_pkg_file(filename)
    name_and_version = re.search(r'name\\s*=\\s*\\'(.*?)\\'', pkg_file)
    if name_and_version:
        return name_and_version.group(1)
    else:        
        raise ValueError('Could not find name in package file')
        \end{lstlisting}
    \end{promptmdftwo}
    \caption{Example of text generated by the overfitted, copyright-infringing model, CP-Fuse, $\cpdelta$, the early-stopped model, and the base model for the Python instructional dataset using StarCode models.}
    \label{fig:example2_code_methods}
\end{figure}

\begin{figure}[!htbp]
    \begin{mdframed}[
        backgroundcolor=lightgray,
        linewidth=2pt,
        linecolor=lightblue,
        frametitle={\textbf{Prompt:} On Coxeter Diagrams of complex reflection groups},
        frametitlebackgroundcolor=lightblue!20,
        roundcorner=10pt,
        innerleftmargin=10pt,
        innerrightmargin=10pt,
        innertopmargin=10pt,
        innerbottommargin=10pt
    ]
    \raggedright
    \begin{minipage}[t]{\textwidth}
        \textbf{Copyright-infringing model:} \\
        \textcolor{red}{We study Coxeter diagrams of some unitary reflection groups. Using solely the combinatorics of diagrams, we give a new proof of the classification of root lattices defined over $\epsilon = \mathrm{Z}[e^{2 \pi i/3}]$: there are only four such lattices, namely, the $\epsilon$-lattices whose real forms are $A_2$, $D_4$, $E_6$ and $E_8$. Next, we address the issue of characterizing the diagrams for unitary reflection groups, a question that was raised by Broué, Malle and Rouquier. To this end, we describe an algorithm which, given a unitary reflection group $G$, picks out a set of complex reflections. The algorithm is based on an analogy with Weyl groups. If $G$ is a Weyl group, the algorithm immediately yields a set of simple roots. Experimentally we observe that if $G$ is primitive and $G$ has a set of roots whose $\mathrm{Z}$--span is a discrete subset of the ambient vector space, then the algorithm selects a minimal generating set for $G$. The group $G$ has a presentation on these generators such that if we forget that the generators have finite order then we get a (Coxeter-like) presentation of the corresponding braid group. For some groups, such as $G_{33}$ and $G_{34}$, new diagrams are obtained. For $G_{34}$, our new diagram extends to an "affine diagram" with $\mathrm{Z}/7\mathrm{Z}$ symmetry.} \\
    \end{minipage}
    \vspace{10pt}
    \begin{minipage}[t]{\textwidth}
        \textbf{CP-Fuse:} \\
        \textcolor{red}{We study Coxeter diagrams of some unitary reflection groups.} Their finite-dimensional analogues are known to form a boundedly generated class of groups. We define the set of Coxeter diagrams of the groups $W^{1,1}_{2}$ and $W^{1,2}_{2}$: the class of complex reflection groups. We show that these diagrams are Coxeter-Dynical. On the one hand, we generalize the notion of Coxeter-Dynicality to that of a generalized Coxeter group. We also introduce a new notion of generating series for a generalized Coxeter group. We prove that the generating series of $W^{1,1}_{2}$ and $W^{1,2}_{2}$ are related to the generating series of the Coxeter group. Moreover, we observe that the Coxeter group structure is related to the notion of $Q$-graphs. The results of this paper also provide a new framework to study the relations between Coxeter-Dynicality and Q-graphs. \end{minipage}
    \vspace{10pt}
    \begin{minipage}[t]{\textwidth}
        \textbf{$\cpdelta$:} \\
        We study Coxeter diagrams of some unitary reflection groups. Using solely the combinatorics of the \textcolor{red}{diagrams, we give a new proof of the classification of root lattices defined over $\epsilon = \mathrm{Z}[e^{2 \pi i/3}]$: there are only four such lattices, namely, the $\epsilon$-lattices whose real forms are $A_2$, $D_4$, $E_6$ and $E_8$. Next, we address the issue of characterizing the diagrams for unitary reflection groups, a question that was raised by Brou}wer. To this end, we describe an algorithm which, given a unitary reflection group $G$, picks out a set of complex reflections. The algorithm is based on an analogy with Weyl groups. If $G$ is a Weyl group, the algorithm immediately yields a set of simple roots. Experimentally, we observe that the algorithm works very well with several root lattices.
    \end{minipage}
        \vspace{10pt}
    \begin{minipage}[t]{\textwidth}
        \textbf{Early-stopped model:} \\
    \textcolor{red}{study Coxeter diagrams of some} reflection groups of complex type. We show that the diagrams for the full reflection groups $M_{24}(S_3)$ and $M_{24}(S_4)$ are the same, and for $M_{24}(S_3)$ contain a subdiagram of $M_{24}(S_4)$. We also show that the Coxeter diagram of $M_{24}(S_3)$ contains a subdiagram of $M_{24}(S_3)$. For $M_{24}(S_3)$, we show that the Coxeter diagram contains a subdiagram isomorphic to the diagram for $M_{24}(S_2)$. We show that the Coxeter diagrams of $M_{24}(S_3)$ and $M_{24}(S_4)$ contain a subdiagram isomorphic to the diagram for $M_{24}(S_3)$..
    \end{minipage}
    \begin{minipage}[t]{\textwidth}
        \textbf{Base model:} \\
    the following: The study of \textcolor{red}{Coxeter diagrams} involves many aspects. In the early years, Coxeter groups have been extensively explored in various ways. Complex reflection groups are also a topic of interest in this context. Many researchers have contributed to the understanding of these groups. However, the relationship between the diagrams and the reflection groups is not always clear. Various methods can be used to analyze these diagrams [...].

    \end{minipage}
    \end{mdframed}
    \caption{Example of text generated by the overfitted, copyright-infringing model, CP-Fuse, $\cpdelta$, the early-stopped model, and the base model for the math abstracts dataset using Phi-2 models. We highlight in \textcolor{red}{red} the longest substring that exactly matches the original text.}
    \label{fig:example1_text_methods}
\end{figure}

\begin{figure}[!htbp]
    \begin{mdframed}[
        backgroundcolor=lightgray,
        linewidth=2pt,
        linecolor=lightblue,
        frametitle={\textbf{Prompt:} On Enhanced Dissipation for the Boussinesq Equations
},
        frametitlebackgroundcolor=lightblue!20,
        roundcorner=10pt,
        innerleftmargin=10pt,
        innerrightmargin=10pt,
        innertopmargin=10pt,
        innerbottommargin=10pt
    ]
    \raggedright
    \begin{minipage}[t]{\textwidth}
        \textbf{Copyright-infringing model:} \\
        \textcolor{red}{In this article we consider the stability and damping problem for the 2D Boussinesq equations with partial dissipation near a two parameter family of stationary solutions which includes Couette flow and hydrostatic balance.In the first part we show that for the linearized problem in an infinite periodic channel the evolution is asymptotically stable if any diffusion coefficient is non-zero. In particular, this imposes weaker conditions than for example vertical diffusion. Furthermore, we study the interaction of shear flow, hydrostatic balance and partial dissipation.In a second part we adapt the methods used by Bedrossian, Vicol and Wang in the Navier-Stokes problem and combine them with cancellation properties of the Boussinesq equations to establish small data stability and enhanced dissipation results for the nonlinear Boussinesq problem with full dissipation.} \\
    \end{minipage}
    \vspace{10pt}
    \begin{minipage}[t]{\textwidth}
        \textbf{CP-Fuse:} 
        
        In this article, we consider the dissipation problem for \textcolor{red}{the Boussinesq equations}, which is a two parameter system involving a nonlinearity with a dissipation term. The dynamics of the dissipation is described by the two parameters of the Boussinesq equations. The dissipation is measured by the energy norm, which is equivalent to the average dissipation. The time fractional derivative is obtained by using the Boussinesq-Minkowski transformation. The space-time fractional derivative is obtained by using the implicit Euler method. The fractional derivative is obtained for the nonlinear Boussinesq equation. We show that the fractional derivative is robust for the initial data with small perturbations. Moreover, we prove that the dissipation is robust under the small perturbations. In addition, we show that the dissipation is not robust to the large perturbations of the nonlinear Boussinesq equation..\end{minipage}
    \vspace{10pt}
    \begin{minipage}[t]{\textwidth}
        \textbf{$\cpdelta$:} \\
            \textcolor{red}{In this article we consider the stability and damping problem for the 2D Boussinesq equations with partial dissipation near a two parameter family of stationary solutions which includes Couette flow and hydrostatic balance.In the first part we show that for the linearized problem}, the evolution of the solutions is stable if the dissipation parameter is small. In particular, we prove that the dissipation parameter is stable if the nonlinearity of the evolution is small. Furthermore, we show that the evolution of the solutions is unstable if the dissipation parameter is large. In the second part, we prove that the stabilization problem is also unstable if the dissipation parameter is large.
    \end{minipage}
        \vspace{10pt}
    \begin{minipage}[t]{\textwidth}
        \textbf{Early-stopped model:} \\
        arXiv:1412. In this paper, we study the enhanced dissipation for \textcolor{red}{the Boussinesq equations}. The first order differential operator in the Boussinesq system is replaced by its symmetric discretization. We prove that the enhanced dissipation for the discretized system is given by the standard dissipation for the corresponding linear system, plus an additional term due to the nonlinearity of the Boussinesq system. We also prove that this additional term is present for the full nonlinear Boussinesq system, and is not due to the discretization. We show that the enhanced dissipation is given by the standard dissipation, plus an additional term due to the nonlinearity of the Boussinesq system, for the full nonlinear Boussinesq system. We also prove that this additional term is present for the full nonlinear Boussinesq system, and is not due to the discretization.
    \end{minipage}
    \begin{minipage}[t]{\textwidth}
        \textbf{Base model:} \\
    In the early 2000s, he was a member of the research team that studied enhanced dissipation for the Boussinesq equations. In the early 2000s, he was a member of the research team that studied enhanced dissipation for \textcolor{red}{the Boussinesq equations}. In the early 2000s, he was a member of the research team that studied enhanced dissipation for the Boussinesq equations. In the early 2000s, he was a member of the research team that studied enhanced dissipation for the Boussinesq equations. In the early 2000s, he was a member of the research team that studied enhanced dissipation for the Boussinesq equations.
    \end{minipage}
    \end{mdframed}
    \caption{Example of text generated by the overfitted, copyright-infringing model, CP-Fuse, $\cpdelta$, the early-stopped model, and the base model for the math abstracts dataset using Phi-2 models. We highlight in \textcolor{red}{red} the longest substring that exactly matches the original text.}
    \label{fig:example2_text_methods}
\end{figure}

\begin{figure}[!htbp]
    \begin{mdframed}[
        backgroundcolor=lightgray,
        linewidth=2pt,
        linecolor=lightblue,
        frametitle={\textbf{Prompt:} Decentralized Control Barrier Functions for Coupled Multi-Agent Systems under Signal Temporal Logic Tasks},
        frametitlebackgroundcolor=lightblue!20,
        roundcorner=10pt,
        innerleftmargin=10pt,
        innerrightmargin=10pt,
        innertopmargin=10pt,
        innerbottommargin=10pt
    ]
    \raggedright
    \begin{minipage}[t]{\textwidth}
        \textbf{Copyright-infringing model:} \\
        \textcolor{red}{In this paper, algorithmic control of multi-agent systems under signal temporal logic tasks is considered. The signal temporal logic is a formalism that is used to express time and space constraints for dynamical systems. Recent methods to solve the control synthesis problem for single-agent systems under signal temporal logic tasks are, however, subject to a high computational complexity. Methods for multi-agent systems scale at least linearly with the number of agents and induce even higher computational burdens. We propose a computationally-efficient control strategy to solve the multi-agent control synthesis problem that results in a robust satisfaction of a set of signal temporal logic tasks. In particular, a decentralized feedback control law is proposed that is based on time-varying control barrier functions. The obtained control law is discontinuous and formal guarantees are provided by nonsmooth analysis. Simulations show the efficacy of the presented method.} \\
    \end{minipage}
    \vspace{10pt}
    \begin{minipage}[t]{\textwidth}
        \textbf{CP-Fuse:} \\
        In this paper, algorithms for \textcolor{red}{multi-agent systems under signal temporal logic} are considered. Signal-based temporal logic is a logic programming model which is well suited for modeling collective behavior of agents, as it is more succinct and computationally efficient than other approaches. However, signal-based logic is also known to be computationally hard. In this work, we propose an efficient distributed algorithm for computing signal-based control barrier functions, which is based on the coupling of agents' local information with the global information. Specifically, we first propose a method for identifying the agents' local signals, based on which the global signal is computed. Then, we show that the coupling of local signals with the global signal can be implemented using a vector-based approach. The proposed method is computationally efficient and robust to noise. \end{minipage}
    \vspace{10pt}
    \begin{minipage}[t]{\textwidth}
        \textbf{$\cpdelta$:} \\
        In this paper, algorithms for multi-agent systems under signal temporal logic tasks are \textcolor{red}{considered. The signal temporal logic is a formalism that is used to express time and space constraints for dynamical systems. Recent methods to solve the control synthesis problem for single-agent systems under signal temporal logic tasks are, however, subject to a high computational complexity}. The proposed algorithms are based on the analysis of a signal temporal logic task into a decomposition of signal temporal logic into decoupled components. Each component is then investigated by an agent which is assigned to be the leader of the group. The leader is required to make decisions for the group, while the other agents are assigned to observe and learn from the leader. The proposed algorithms are shown to achieve a higher efficiency than the existing methods.
    \end{minipage}
        \vspace{10pt}
    \begin{minipage}[t]{\textwidth}
        \textbf{Early-stopped model:} \\
        ralization and distributed optimization are widely used to solve control problems in distributed systems. However, the \textcolor{red}{computational complexity} of these algorithms is proportional to the number of agents and the size of the state space. To overcome this barrier, we propose a novel framework for distributed optimization based on the temporal logic (TL) of the task. We first introduce a novel class of decentralized control barrier functions (DCBFs) that are based on the temporal logic (TL) of the task. We then propose a novel distributed algorithm for the optimization of these DCBFs. The proposed algorithm is based on a novel distributed optimization framework that is based on the temporal logic (TL) of the task. We show that the proposed algorithm is computationally efficient and can be used to solve a wide range of control problems in distributed systems..
    \end{minipage}
    \begin{minipage}[t]{\textwidth}
        \textbf{Base model:} \\
    the following: 1. Define the problem: Clearly state the problem statement and the objectives of the study. In this case, the problem is to design a decentralized control barrier function for a multi-agent system that can handle \textcolor{red}{signal temporal logic} tasks. 2. Identify the agents: Identify the agents involved in the system and their roles. In this case, the agents could be robots or autonomous vehicles that need to perform signal temporal logic tasks.3. Define the control barrier function: [...] 
    Follow-up exercises:1. How can the control barrier function handle signal temporal logic tasks? Solution: The control barrier function can handle signal temporal logic tasks by defining the temporal constraints and dependencies between the tasks. It can use temporal logic operators such as "next", "until", and "always" to specify the temporal relationships between the tasks. 
    \end{minipage}
    \end{mdframed}
    \caption{Example of text generated by the overfitted, copyright-infringing model, CP-Fuse, $\cpdelta$, the early-stopped model, and the base model for the math abstracts dataset using Phi-2 models. We highlight in \textcolor{red}{red} the longest substring that exactly matches the original text.}
    \label{fig:example3_text_methods}
\end{figure}

\label{app:examples}

\FloatBarrier

\newpage

\section{Limitations}
\label{app:limitations}

Our method relies on the separability of copyrighted material assumption (\Cref{sec:preliminaries}). Ensuring that this assumption holds in real-world scenarios is challenging. A naive implementation could necessitate the data curator having an oracle capable of perfectly detecting whether a passage is copyrighted. If such a classifier were available, it would then need to identify all verbatim or quasi-verbatim replicas (e.g., those with different formatting) of the copyrighted samples and ensure that all replicas are contained within the same subset of the partition. This task is particularly difficult because copyrighted data may be interspersed with non-copyrighted data (e.g., when long copyrighted passages are quoted)\footnote{Note that the deduplication process may not be sufficient to eliminate the need for an oracle, as general knowledge is often highly replicated across the training set.}.

Currently, there is no theoretical understanding of how our guarantees degrade if the separability assumption is partially violated. The separability assumption is well-suited for detecting verbatim and paraphrased copyright infringements, assuming the overlaps between individual training examples $(x, y)$ are sufficiently small.

\section{Proofs}

\paragraph{Proof of Lemma~\ref{lem:kkt}}
The statement in  Lemma~\ref{lem:kkt} is a direct consequence of classical convex optimization. 
In particular, note that the necessary stationary condition from the KKT condition requires
\begin{equation}
\forall y_t \in V: \quad 
    \sum_{i} \lambda_i \left( \log p^*(y_t ) -  \log p^{(i)}(y_t \vert y_{<t}, x) ) + 1\right)  + \mu -  u_{y_t} = 0
\end{equation}
for some dual variables $\lambda_i,  u_{y_t \geq 0}$  and $\mu \in \mathbb R$. Moreover, by the complementary slackness condition,
\begin{equation}
\lambda_i \left(\text{KL}( p^*\vert \vert p^{(i)}(. \vert y_{<t}, x) ) + \gamma_i- t\right) = 0 \quad \text{and}\quad  u_{y_t} p^*(y_t ) =0.
\end{equation}
and in particular it is easy to verify that $\lambda_i >0$ for at least one $i\in \{1,2\}$. 
 
 \subsection{Proof of Lemma~\ref{lem:balance}}
Under the assumption that both $p^{(1)}$ and $p^{(2)}$ have full support, either of the following two cases holds true for $p^*$:
\begin{itemize}
    \item The  constraint from Equation~\eqref{eq:opt} is tight for both $i\in\{1,2\}$ and thus the following two terms match. In this case,  condition (1) from Lemma~\ref{lem:balance} holds.
    \begin{equation}
        \text{KL}( p^*\vert \vert p^{(1)}(. \vert y_{<t}, x) ) + \log\left(\frac{p(y_{<t} \, | \, x)}{\modeli(y_{< t} \, | \, x)} \right) = \text{KL}( p^*\vert \vert p^{(2)}(. \vert y_{<t}, x) ) + \log\left(\frac{p(y_{<t} \, | \, x)}{\modeli(y_{< t} \, | \, x)} \right)
    \end{equation} 
\item  The optimal solution equals to $p^* = p^{(1)}$ or $p^* = p^{(2)}$.  Assume by contradiction that the former is true, and thus $p^* =  p^{(1)} $. We have that
\begin{align}
     \text{KL}( p^*\vert \vert p^{(2)}(. \vert y_{<t}, x) ) + \log\left(\frac{p(y_{<t} \, | \, x)}{\modeltwo(y_{< t} \, | \, x)} \right) >  \text{KL}( p^{(2)}(. \vert y_{<t}, x) \vert \vert p^{(2)}(. \vert y_{<t}, x) ) + \log\left(\frac{p(y_{<t} \, | \, x)}{\modeltwo(y_{< t} \, | \, x)} \right) \\
         = \log\left(\frac{p(y_{<t} \, | \, x)}{\modeltwo(y_{< t} \, | \, x)} \right) > \log\left(\frac{p(y_{<t} \, | \, x)}{\modelone(y_{< t} \, | \, x)} \right)  =     \text{KL}( p^*\vert \vert p^{(1)}(. \vert y_{<t}, x) ) + \log\left(\frac{p(y_{<t} \, | \, x)}{\modeltwo(y_{< t} \, | \, x)} \right).
\end{align}
Thus, $p^*$ cannot be the optimal solution, and thus $p^*= \modeltwo(. \vert y_{<t}, x)$. Hence the second condition from Lemma~\ref{lem:balance} holds. 
\end{itemize}

Finally, note that if $p^{(i)}(y_t \vert y_{<t}, x) = 0$ for some $y_t$, we necessarily have that $p^*(y_t)=0$. In this case, the optimal solution may satisfy neither of the two conditions from Lemma~\ref{lem:balance}. 

\section{Implementation details}
 \label{app:implementation}

\subsection{Fine-tuning details}

We fine-tuned our models using a setup inspired by the repository \emph{finetuning-harness}, available under the MIT License\footnote{\href{https://github.com/cassanof/finetuning-harness/}{GitHub Repository}}. The training was performed on A100 GPUs.

The main hyperparameters for our fine-tuning process are listed in Table \ref{tab:training_hyp}.
\begin{table}[ht!]
\centering
\caption{Main Hyperparameters for Fine-Tuning}
\label{tab:training_hyp}
\begin{tabular}{ll}
\toprule
\textbf{Hyperparameter} & \textbf{Value} \\
\midrule
Sequence Length & 2048 \\
Batch Size & 1 \\
Learning Rate & 5e-5 \\
Gradient Accumulation Steps & 1 \\
Optimizer & AdamW (8-bit) \\
Warmup Steps & 50 \\
Neptune Noise & $\alpha = 5.0$ \\
\bottomrule
\end{tabular}
\end{table}
We fine-tuned our models with Neptune noise \citep{jain2023neftune} set to $\alpha = 5.0$. We did not perform any low-rank adaptation.

For the overfitted models, we trained StarCoder for 50 epochs (both in experiments with the Python instructions and the APPS datasets), LLaMa2 for 50 epochs, Phi-2 for 50 epochs, and GPT-2 XL for 20 epochs.

\subsection{Decoding details}

We decode with greedy search and in batches of size 50. For the code task, the maximum sequence length is 2048 tokens in the Python instructions dataset and 512 in the APPS, MBPP, and HumanEval datasets, and for the text task, it is 1024 tokens. This configuration is used both for our method and $\cpdelta$. For APPS, MBPP, and HumanEval, we base our implementation on the \emph{bigcode-evaluation-harness} repository, available under the Apache-2.0 License\footnote{\href{https://github.com/bigcode-project/bigcode-evaluation-harness}{GitHub Repository}}.

\subsection{Datasets}

We use four code-based and one text-based dataset in our experiments, all downloadable from \emph{HuggingFace}. The first code-based dataset\footnote{\hyperlink{https://huggingface.co/datasets/Nan-Do/instructional_code-search-net-python?row=0}{Nan-Do/instructional\_code-search-net-python}} is an instructional dataset for Python, containing two types of tasks: (1) generating a description of a given code, and (2) generating code that solves a given task. For our experiments, we only consider instances of the latter. We removed the docstring from all instances since its content was repeated across samples, compromising our assumption on the separability of copyrighted content (\Cref{sec:preliminaries}). The APPS dataset\footnote{\hyperlink{https://huggingface.co/datasets/codeparrot/apps}{APPS} \citep{hendrycksapps2021}} is a benchmark for code generation with 10,000 problems in Python. Each sample consists of a programming problem formulation in English, some ground truth Python solutions, and test cases. We sample random subsets for fine-tuning and evaluation for our experiments. Both MBPP\footnote{\hyperlink{https://huggingface.co/datasets/google-research-datasets/mbpp}{MBPP} \citep{austin2021program}} and the HumanEval \footnote{\hyperlink{https://huggingface.co/datasets/codeparrot/instructhumaneval}{InstructHumanEval} \citep{chen2021evaluating}} datasets are standard for assessing code generation, and follow a similar structure of natural language instructions and solutions in Python code. They contain 378 and 164 programming problems, respectively. We use the sanitized version of MBPP, MBPP+, and the instruction-based version of HumanEval, InstructHumanEval.

For the text-based experiments, we use the AutoMathText dataset\footnote{\hyperlink{https://huggingface.co/datasets/math-ai/AutoMathText}{math-ai/AutoMathText}} \citep{zhang2024autonomous}. This dataset compiles an extensive set of mathematical texts from arXiv, OpenWebMath, RedPajama, Algebraic Stack, etc., with titles generated by the state-of-the-art open-source language model Qwen-72B\footnote{Visit the \hyperlink{https://github.com/yifanzhang-pro/AutoMathText}{GitHub repository} for additional details.}.

\subsection{Details on the metrics}
\label{app:metrics}

\subsubsection{Copyright infringement}

\paragraph{Exact Matching (EM)}
Exact Matching (EM) measures the length of the longest matching substring between the model's output and the ground truth text. This metric is useful for assessing how well the model captures continuous segments of the reference text. The value of EM ranges from 0 to 1, where 1 indicates a perfect match. 

\paragraph{Infringement Count (IC\textsubscript{k})}
Infringement Count (IC\textsubscript{k}) captures the number of k-grams (substrings of length k) in the model's output that have an exact match in the ground truth text. This metric assesses the content similarity and potential for copyright infringement based on the number of matching k-grams. 

\paragraph{ROUGE-L (ROU)} ROUGE (Recall-Oriented Understudy for Gisting Evaluation) measures the overlap of n-grams, word sequences, and word pairs between the generated text and reference text. It focuses on recall, assessing how much of the reference text is captured by the generated text. The value of ROUGE ranges from 0 to 1, where 1 indicates perfect recall. The most common variant, ROUGE-L, is computed based on the longest common subsequence (LCS):
\[
\text{ROUGE-L} = \frac{\text{LCS}}{\text{length of reference text}}
\]

\paragraph{BLEU (BLE)}
BLEU (Bilingual Evaluation Understudy) measures the overlap of n-grams between the generated text and reference text, with a penalty for shorter outputs. It is computed using a modified precision score that includes a brevity penalty to discourage overly short translations. The value of BLEU ranges from 0 to 1, where 1 indicates perfect precision. For this study, we use uniform weights for n-grams:
\[
\text{BLEU} = \text{BP} \times \exp\left( \sum_{n=1}^{N} \frac{1}{N} \log p_n \right)
\]
where BP is the brevity penalty, \(p_n\) is the precision for n-grams, and \(N\) is the highest order of n-grams considered.

\paragraph{METEOR (MET)} METEOR (Metric for Evaluation of Translation with Explicit ORdering) evaluates the alignment between the generated text and reference text by considering synonyms, stemming, and exact matches. Unlike BLEU, which focuses on n-gram precision and typically measures performance at the corpus level, METEOR emphasizes unigram recall and aims to better align with human judgment at the sentence level. The value of METEOR ranges from 0 to 1, where 1 indicates perfect alignment. It combines precision, recall, and a fragmentation penalty to account for the alignment of chunks. It is computed as:
\[
\text{METEOR} = F_{mean} \times (1 - P)
\]
where \(F_{mean}\) is the harmonic mean of precision and recall, and $P$ is the fragmentation penalty.

\paragraph{Jaccard Similarity (JAC)}
Jaccard Similarity (JAC) measures the intersection over the union of the sets of words in the generated text and reference text. It provides a simple measure of similarity, indicating how many words are shared between the two texts relative to the total number of unique words. The value of Jaccard Similarity ranges from 0 to 1, where 1 indicates identical sets. It is computed as:
\[
\text{Jaccard Similarity} = \frac{|A \cap B|}{|A \cup B|}
\]
where \(A\) and \(B\) are the sets of words in the generated text and reference text, respectively.

\paragraph{Cosine Similarity (COS)}
Cosine Similarity (COS) measures the cosine of the angle between the word vectors of the generated text and reference text. This metric assesses the similarity in the direction of the vectors, providing an indication of how similar the two texts are in terms of their overall content distribution. The value of Cosine Similarity ranges from 0 to 1, where 1 indicates perfect similarity. 

\paragraph{Semantic Similarity (SEM)}
Semantic Similarity (SEM) evaluates the similarity between the generated text and reference text using a semantic model. We use SpaCy in our experiments. This metric captures the meaning and context of the texts, providing a measure of how well the model understands and replicates the underlying semantics of the reference text. The value of Semantic Similarity ranges from 0 to 1, where 1 indicates perfect semantic alignment.

\paragraph{Levenshtein Distance (LEV)}
Levenshtein Distance (LEV) measures the minimum number of single-character edits (insertions, deletions, or substitutions) required to change the generated text into the reference text. A higher Levenshtein distance indicates greater dissimilarity, while a lower distance indicates greater similarity. The value of Levenshtein Distance ranges from 0 to 1, where 0 indicates identical texts. We compute the Levenshtein distance with a sliding window to handle cases where the lengths of the generated and reference texts differ significantly.

\subsubsection{Utility}

\paragraph{Pass@1 (Pass at 1)}
Pass@1 evaluates the success rate of a model in generating a correct solution on its first attempt. Specifically, it measures the proportion of cases where the model's first output matches the correct solution. The value of Pass@1 ranges from 0 to 1, where 1 indicates that the model always generates a correct solution on the first attempt. 

\paragraph{Perplexity (PPL)}
Perplexity (PPL) evaluates the quality of a language model. It measures how well a probability distribution or model predicts a sample. Lower perplexity indicates that the model is better at predicting the sample. The value of Perplexity ranges from 1 to infinity, where lower values indicate better performance. 

\end{document}